\documentclass[twoside,11pt]{article}
\usepackage{jmlr2e}
\usepackage[format=hang]{caption}

\usepackage[toc, acronym, nogroupskip, nonumberlist, nopostdot]{glossaries}
\usepackage{algorithm}
\usepackage{algpseudocode}
\usepackage{graphicx}
\usepackage{subcaption}
\usepackage{dcolumn}% Align table columns on decimal point
\usepackage{bm}% bold math
\usepackage[dvipsnames]{xcolor}
\usepackage{booktabs} % better tables

\newacronym{sa}{SA}{Simulated Annealing}
\newacronym{lahc}{LAHC}{Late-Acceptance Hill Climbing}
\newacronym{fbn}{FBN}{Feedforward Boolean Network}
\newacronym{cl}{CL}{Curriculum Learning}
\newacronym{spl}{SPL}{Self-Paced Learning}
\newacronym{mtl}{MTL}{Multi Task Learning}
\newacronym{mlc}{MLC}{Multi Label Classification}
\newacronym{dag}{DAG}{directed acyclic graph}
\newacronym{kfs}{kFS}{k-Feature-Set}
\newacronym{mfs}{minFS}{Minimum-Feature-Set}
\newacronym{grn}{GRN}{Gene Regulatory Network}
\makeglossaries

\jmlrheading{18}{2017}{1-26}{1/17; Revised 5/17}{11/17}{17-007}{Shannon Fenn and Pablo Moscato}

\ShortHeadings{Target Curricula via Selection of Minimum Feature Sets}{Fenn and Moscato}
\firstpageno{1}

\begin{document}

\title{
Target Curricula via Selection of Minimum Feature Sets:\\a Case Study in Boolean Networks
}

\author{\name Shannon Fenn \email shannon.fenn@newcastle.edu.au \AND
       \name Pablo Moscato \email pablo.moscato@newcastle.edu.au \\
       \addr School of Electrical Engineering and Computing\\
       University of Newcastle \\
       University Drive \\ Callaghan NSW 2308 \\ Australia}

\editor{Amos Storkey}

\maketitle

\begin{abstract}%   <- trailing '%' for backward compatibility of .sty file

We consider the effect of introducing a curriculum of targets when training Boolean models on supervised \gls{mlc} problems. In particular, we consider how to order targets in the absence of prior knowledge, and how such a curriculum may be enforced when using meta-heuristics to train discrete non-linear models.

We show that hierarchical dependencies between targets can be exploited by enforcing an appropriate curriculum using hierarchical loss functions. On several multi-output circuit-inference problems with known target difficulties, \glspl{fbn} trained with such a loss function achieve significantly lower out-of-sample error, up to $10\%$ in some cases. This improvement increases as the loss places more emphasis on target order and is strongly correlated with an easy-to-hard curricula. We also demonstrate the same improvements on three real-world models and two \gls{grn} inference problems.

We posit a simple a-priori method for identifying an appropriate target order and estimating the strength of target relationships in Boolean \glspl{mlc}. These methods use \emph{intrinsic dimension} as a proxy for target difficulty, which is estimated using optimal solutions to a combinatorial optimisation problem known as the \gls{mfs} problem. We also demonstrate that the same generalisation gains can be achieved without providing any knowledge of target difficulty.
\end{abstract}


%\tableofcontents

\section{\label{sec:intro}Introduction}

When students are first taught the concept of addition, they are not simply handed an eclectic set of many digit number pairs and their summations; rather, we provide them a learning curve of carefully chosen examples which increase in difficulty (i.e. number of digits). From standard classroom practices we can glean two key sources of inspiration regarding effective learning and teaching processes of humans: the curriculum of examples, and the curriculum of targets. This paper examines the latter.

\cite{khan_how_2011} found that humans naturally follow an easy-to-hard example-based curriculum when teaching a single-target concept to an unknown learner. Methods of scheduling examples this way have been explored in the machine learning community for some time, with positive results for non-convex scenarios~\citep{bengio_curriculum_2009-1}.

The typical explanation for the success of example curricula in non-convex scenarios is that it imposes a form of transfer learning where the simpler concept is discovered first and subsequently informs the more complex concept~\citep{bengio_curriculum_2009-1}. This leads neatly into a consideration of the second inspiration from the classroom: target curricula.

We consider the problem of learning discrete models for \gls{mlc} problems with Boolean inputs, where closed-form solutions and gradient-guided optimisation techniques are unavailable. We use \acrfullpl{fbn}---defined in Section~\ref{sec:bn}---as a case study for implementing target-curriculum regularisation with simple modifications to the typical $L_1$/$L_2$-Loss. We also introduce methods for estimating target overlap and target difficulty order in Boolean \glspl{mlc} by finding a lower bound to the intrinsic dimension of the induced function. The intrinsic dimension has been presented as a proxy for the overall complexity of arbitrary data sets~\citep{granata_accurate_2016} suggesting its promise as a metric for ordering targets.

\subsection{\label{sec:contrib}Contributions}
We provide three novel contributions:
\begin{itemize}
\item we show that using hierarchical loss functions improves generalisation performance of \glspl{fbn} trained on several circuit-inference problems possessing a natural target progression,
\item we describe a simple method for estimating the complexity and overlap of Boolean targets by lower bounding their intrinsic dimensions using solutions to the \gls{mfs} problem, and
\item we demonstrate the success of the combined method on the same problem instances when knowledge of the target difficulty is withheld.
\end{itemize}

In the remainder of this section we define \glspl{fbn} as a learning model, give background on \gls{mlc} and outline existing work on sample and target curricula. In Section~\ref{sec:order} we define hierarchical loss functions as well as the \gls{mfs} problem and how it can be used to detect target overlap and estimate an appropriate curriculum. Finally, in Section~\ref{sec:experimental} we present results on several problems.

\subsection{\label{sec:bn}Boolean Networks}

This paper focuses on Boolean \gls{mlc} problems where the targets possess differing complexities. In this section we provide definitions of the problem domain, its relevance and the representation of interest to this work: the \acrlong{fbn}.

A Boolean \gls{mlc} problem is one where inputs and targets take values in $\mathcal{B} \in \{0, 1\}$. An intuitive model for such a problem is a digital circuit, or \gls{fbn}. A \gls{fbn} with $l$ inputs and $m$ outputs computes a function of the form $f: \mathcal{B}^{l} \rightarrow \mathcal{B}^{m}$ by chaining computations from internal nodes (commonly called gates).

\begin{figure}
\centering
\includegraphics[width=0.9\textwidth]{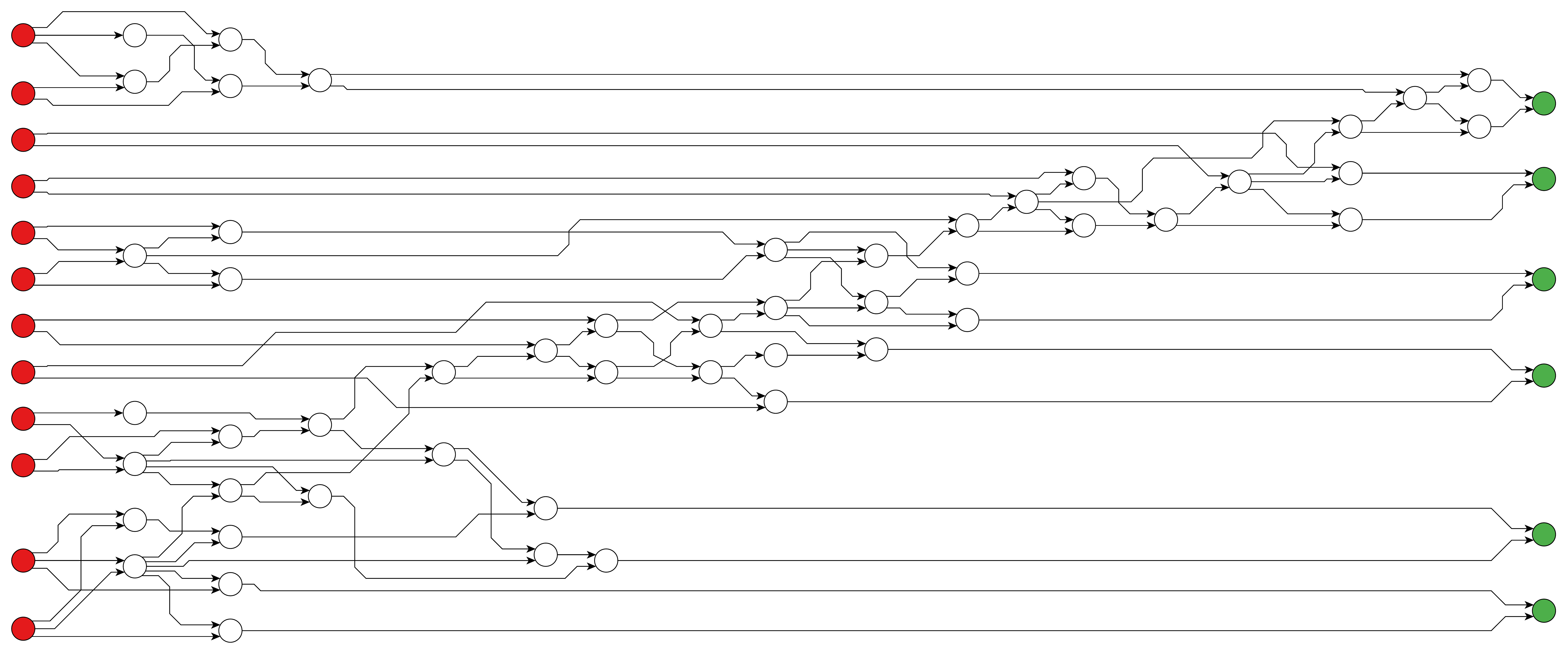}
\caption{A $56$-node \gls{fbn} resulting from our learning procedure, which correctly implements the $6$-bit addition function. Each node takes $2$ inputs and computes the NAND function as its output. Inputs (far left) have been coloured red and outputs (far right) green. Note the ripple-carry style flow of information between sub-networks.}
\label{fig:example_fbn}
\end{figure}

Internally, a \gls{fbn} is a \gls{dag} where each node has an associated value in $\mathcal{B}$. Nodes take the values of their predecessors as input to a Boolean function which they provide as output. Input nodes are those with no predecessors and their value is provided from outside the network. The output of the network is simply the values at a particular subset of externally visible output nodes. A \gls{fbn} can thus compute a function from many inputs to many outputs by combining simpler functional elements. Figure~\ref{fig:example_fbn} shows an example \gls{fbn} which correctly implements a $6$-bit adder.

One advantage to working with this representation is that two singleton functionally complete operator sets exist: the negated-and (NAND) and negated-or (NOR) functions. Every Boolean mapping can be computed by some degree-$2$ \gls{dag} consisting purely of NAND nodes. We will use this representation for the remainder of the paper.

\glspl{fbn} have a number of other advantages as learning models. They are efficient in time and memory to simulate and trivial for direct hardware implementation and logical rule conversion. Logical relationships are simple for humans to extract information from which is important if the goal of the learning procedure is knowledge-generation. While other representations have received significantly more attention, \glspl{fbn} are trivial to map to existing digital hardware---an open problem for neural nets. They can use packed integer representations and bitwise arithmetic to improve model evaluation speed during training and deployment. Recent literature presents methods for binarising the internal layers of deep neural nets precisely for these reasons~\citep{courbariaux_binarized_2016}.

As the name suggests, \glspl{fbn} compute with purely Boolean logic, rather than using weighted sums and continuous activations. This brings challenges, the most prevalent being the lack of gradient-based optimisation procedures or closed form solutions. All optimisation techniques discussed below use general purpose combinatorial meta-heuristics, such as \gls{sa}. Herein all networks were optimised using local search in the space of graph structures---possible due to the functional completeness of the NAND operator.

\subsection{\label{sec:learning_in_with_FBNs}Learning with Boolean Networks}

Since the late publication of Turing's first description of a machine which learns by changing the connections of a NAND-gate circuit~\citep{turing_intelligent_1948} the majority of the work on learning with a Boolean Network model has been done in the field of physics. Some time later, \cite{patarnello_learning_1987}, and subsequently others~\citep{vandenbroeck_learning_1990,goudarzi_learning_2014}, demonstrated that \Glspl{fbn} could generalise well on some problems, despite training set sizes significantly smaller than the space of all possible patterns.

Learning \glspl{fbn} is a \emph{combinatorial} optimisation problem, typically solved using meta-heuristics such as \gls{sa} or genetic algorithms. \cite{patarnello_learning_1987} optimised networks by \gls{sa} in the space of feedforward structures. Each move in the search procedure consisted of a random change in a single connection. Moves which also changed the node activation have been considered~\citep{vandenbroeck_learning_1990, goudarzi_learning_2014}, however in light of the NAND function's functional completeness, this becomes unnecessary.

To simplify the following loss function definitions we define the $n \times m$ error matrix $E$, which fully characterises the mistakes made by a candidate network on a set of $n$ example patterns with $m$ targets. $E$ is the element-wise absolute difference between the network output matrix $Y'$ and the target matrix $Y$, with elements given by
\begin{equation*}
E_{i,j} = \left|Y_{i,j} - Y'_{i,j}\right|,
\end{equation*}
where $E_{i, j} \in \mathcal{B}$ is the error for the $j^{th}$ target for the $i^{th}$ example.

Even for problems with multiple outputs, the only loss functions used to date~\citep{patarnello_learning_1987,teuscher_learning_2007,goudarzi_learning_2014} have been analogues of the $L_1$-loss:\footnote{With purely binary predictors and targets the $L_1$- and $L_2$-loss are identical.}
\begin{equation*}
L_1 = \frac{1}{m \left|\mathcal{I}\right|}\sum_{j=1}^{m} \; \sum_{i\in \mathcal{I}} E_{i, j},
\end{equation*}
where $\mathcal{I}$ is the set of example indices shown to the network. This loss treats all examples and targets equally and is the natural first choice for a guiding function, but for multi-target problems there are other options which we explore in Section~\ref{sec:funcs}.

So far no work has considered differences between learning single and multiple output \glspl{fbn}. Results for \gls{mtl} suggest that, with multiple related target, some networks may be favoured due to target correlation, increasing generalisation performance. To explore the possibility of further improvements, we use \glspl{fbn} as a case study in the effects of target curriculum enforcement by hierarchical loss-functions on purely discrete models. In the following section we outline the relevant work on target curricula.

\subsection{\label{sec:intro-example-curr}Example Curricula}
The approach we present in this work draws strong inspiration from \emph{example} curricula. In this subsection we give a brief outline of some relevant work to better frame our approach.

Curriculum learning by applying a ranking of \emph{examples} can be split into two main camps: \acrfull{cl} and \acrfull{spl}~\citep{kumar_self-paced_2010}. In the former the ranking is determined a priori while in the latter it is learned jointly with the model. \cite{jiang_self-paced_2015} demonstrated a combined approach---intuitively called Self-Paced Curriculum Learning---where constraints on the example-ranking are imposed by a fixed curriculum and the ranking is learned jointly with the model but subject to the curriculum constraints. SPL and CL emerge as special cases of their framework. They present positive results in comparison to baseline methods on matrix factorization and multimedia event detection problems.

\cite{spitkovsky_baby_2009} demonstrated the success of a curriculum of samples of \emph{increasing complexity} on the problem of determining grammar from free-form sentences. Their proxy for complexity in this case is the length of the example sentence. They also noted the appearance of a ``sweet-spot'' in sentence length, the inclusion of samples above which reduced performance. The combination of both their curriculum and complexity limitation improved on the state-of-the-art.

It is important to note that the terms \gls{cl} and \gls{spl} used within this subsection related \emph{example} curricula. It is possible to also consider a learner---in a \gls{mlc} framework---to be following a curriculum of \emph{targets} (or to be self-paced in the same respect). We primarily cover the former case of a fixed curriculum of targets, and throughout this paper ``curriculum'' will refer to a curriculum of targets.

\subsection{\label{sec:intro-target-curr}Target Curricula}
Learning multiple targets concurrently with a shared representation has been shown, theoretically and experimentally, to improve generalisation results when there are relationships between the targets~\citep{caruana_multitask_1997, wang_multiplicative_2016}. The number of publications discussing multi-label and multi-task learning is growing rapidly and using a \emph{curriculum} of targets, while less general, is also beginning to garner attention.

The importance of sharing knowledge between targets increases when there is a disparity in target complexity. \cite{gulcehre_knowledge_2016} showed a visual task, on which many popular methods---including deep neural nets, SVMs and decision trees---failed. This same task became solvable when a simpler hinting target was provided \citep[in much the same way as introduced by][]{abu-mostafa_learning_1990}. However, unless relative target difficulties are known (or suspected), we also face the issue of determining an appropriate order during learning.

\cite{pentina_curriculum_2015} presented a self-paced method using adaptive linear SVMs. The implementation was model-specific, providing information transfer by using the optimal weight vector for the  most recently learned task to regularise the next task's weight vector. They discover the curricula by solving all remaining tasks and selecting the best fitting, resulting in a quadratic increase in training time over solving tasks individually. While this may be acceptable for a linear model, it would be prohibitive for a \gls{fbn}.

~\cite{li_self-paced_2017} implemented task curricula with linear models by weighting the contribution of each task-example pair to the loss function. They used a regulariser in which task-example pairs with low training error receive larger values for these weights. However, this approach does not appear applicable to inferring highly non-linear discrete models.

\cite{lad_toward_2009} present the only model-agnostic approach to determine an optimal order of tasks that we are aware of. They use conditional entropy to produce a pairwise order matrix and solve the resulting NP-Hard Linear Ordering Problem~\citep[see][for a definition]{ceberio_linear_2015}. Their approach is not applicable to Boolean \glspl{mlc} since the conditional entropy difference calculation they use reduces to the difference of the entropies of the individual targets; which, for balanced binary targets, rapidly approaches zero as more training samples are given.

The impact of target order is highlighted by~\cite{read_classifier_2011}. Their approach uses \emph{classifier chains}: independent learners for each binary label, trained in an arbitrary order, with the output of all prior classifiers included as input to each subsequent classifier. The authors note that the ordering of targets has a notable effect on classifier performance; which they address using ensembles of chains, each with a different random order. They note consistent superiority of the ensemble approach over a single random ordering, which suggests non-trivial variability in the impact of different orderings. Due to the large space of permutations, a principled approach to determine appropriate orderings a priori would be of value.

We have not seen work using hierarchical loss-functions (or similar regularisers) to impose a target curriculum, nor have we seen methods for determining a curriculum by use of feature selection methods. The following section describes our approach.

\section{\label{sec:order}Learning Targets in Order}

Here we describe three loss functions which we used to enforce a curriculum of targets, as well as a novel method for obtaining said curriculum which reduces the problem to $m$ instances of the \acrfull{mfs} problem.

\subsection{\label{sec:funcs}Hierarchical Loss Functions}

Assume a two-target Boolean \gls{mlc} for which we have three candidate learners, and assume we know that the first target is ``easier'' under some definition than the second. The first learner labels target $1$ correctly and target $2$ incorrectly for all examples. The second learner does the opposite, and the third learner labels $50\%$ of the examples incorrectly for both targets. All three networks will have an equal $L1$-loss.

The rationale behind the loss functions we are about to define is that the first of the above learners should be preferred. By focusing on easier targets earlier in the training process the network is more likely to find substructures useful to later targets. The expectation is faster convergence speed, better generalisation or both.

The $L_1$-loss is simply the mean of the error matrix. In addition, we consider three loss functions which hierarchically aggregate the error matrix elements in progressively more strict ways:
\begin{itemize}
\item a linearly weighted mean: $L_{w}$,
\item a locally hierarchical mean: $L_{lh}$, and
\item a globally hierarchical mean: $L_{gh}$.
\end{itemize}
All definitions below assume the targets are ordered by difficulty, with the simplest target at the lowest index.

The first function, $L_w$, encourages the target curriculum by weighting the easier targets more highly. The weighting is linear and the resulting summation is normalised to the unit interval:
\begin{equation*}
L_{w} = \frac{2}{m\left(m - 1\right)\left|\mathcal{I}\right|}\sum_{i\in \mathcal{I}} \sum_{j=1}^{m} \; \left(m-j+1\right) E_{i,j} \text{ .}
\end{equation*}

We named the second function, $L_{lh}$, the \emph{locally hierarchical} loss, as it enforces a hierarchy of targets on a per-example basis. The learner is rewarded for correctly labelling target $j$ of example $i$ if and only if it has also correctly labelled all prior targets $\left\lbrace 1, 2, ..., j-1\right\rbrace$ \emph{for that example}. To define this function, for each example~$i$, we define the recurrence:
\begin{align*}
% a_{i,1} &= E_{i,1} \\
% a_{i,k} &= a_{i,k-1} \cdot \left(1 - E_{i,k}\right) + E_{i,k} \text{ ,}
a_{i,1} &= E_{i,1} \\
a_{i,k+1} &=
\begin{cases}
  E_{i, k+1}, & \text{if $a_{i, k} = 0$} \\
    1, & \text{if $a_{i, k} = 1$}
\end{cases}\text{ ,}
\end{align*}
from which we can define
\begin{equation*}
L_{lh} = \frac{1}{m\left|\mathcal{I}\right|}\sum_{i\in \mathcal{I}}\sum_{k=1}^{m} a_{i, k} \text{ .}
\end{equation*}

The final loss, $L_{gh}$, follows the same principle as the former but across all examples. For this reason we named it the \emph{globally hierarchical} loss. In this case the learner is given reward for correctly labelling target $j$ (of any example) if and only if it has also correctly labelled all prior targets $\left\lbrace 1, 2, ..., j-1\right\rbrace$ for all examples given. This function is effectively a ``soft'' equivalent to learning the targets rigidly in order. We can define $L_{gh}$ using the mean per-target errors (the row-means of $E$):
\begin{equation*}
\delta_{j} = \frac{1}{\left|\mathcal{I}\right|}\sum_{i\in \mathcal{I}} E_{i, j} \text{ .}
\end{equation*}
These can be used to define another recurrence,
\begin{align*}
b_{0} &= \delta_{0} \\
b_{k+1} &=
\begin{cases}
  \delta_{k+1}, & \text{if $b_{k} = 0$} \\
    1, & \text{if $b_{k} > 0$}
\end{cases}\text{ ,}
\end{align*}
giving
\begin{equation*}
L_{gh} = \frac{1}{m}\sum_{j=1}^{m} b_j \text{ .}
\end{equation*}

Unless the difficulty order is known, or suspected, from domain specific knowledge we also require a method for estimating the target order from the training examples. We describe a method suitable for Boolean or categorical \glspl{mlc} in the following section.

\subsection{\label{sec:ordering-top}Measuring Target Overlap and Difficulty Order}

The key requirement for ordering targets is an estimation of their relative difficulty. It is likely most suitable to use domain knowledge where available, however in cases where this is impossible or difficult to obtain we suggest a rule of thumb: order by increasing intrinsic dimension. We present an intuitive approach below using a combinatorial problem which did not significantly increase the learning time for the instances considered. In Section~\ref{sec:eval_order} we show that the target order estimated by this method gives generalisation improvements close to those obtained using the known order (Section~\ref{sec:lf_results}).

Our key assumption is that a target can be expected to be simpler than another if its intrinsic dimension (the number of input variables on which it truly depends) is smaller. This assumption is strong but there is also a strong basis for it as a rule-of-thumb in the Boolean domain, as a function's intrinsic dimension places an upper bound on the size of the smallest \gls{fbn} which implements it. Furthermore, intrinsic dimension is showing promise as a method for estimating the complexity of arbitrary data~\citep{granata_accurate_2016} and as such we expect the validity of this idea may extend beyond the Boolean case.

The likelihood of effective information transfer can also be estimated by measuring the overlap in intrinsic input features. Thus estimating both target overlap, and relative complexity, is reduced to the problem of determining the minimum set of input features on which a valid function can be defined; that is, where no pattern is mapped to multiple target values. This is well known as the \acrfull{mfs} problem.

\subsubsection{\label{sec:minfs}The Minimum-Feature-Set Problem}

The smallest intrinsic dimension of any Boolean function which is in full agreement with given set of examples can be found using the \acrfull{mfs} problem. It asks: what is the smallest subset of given input features, on which a single target can still form a non-contradictory mapping? This problem, and particularly this view of it, is not in any way new to the machine learning community~\citep{davies_np-completeness_1994}. However, to our knowledge, it has not been used to estimate relative target difficulty before.

This equivalent decision problem is known as the \emph{\acrfull{kfs}} problem. Formally:

\emph{\acrlong{kfs}}

Given: A binary $n\times p$ matrix, $\mathcal{M}$, where the rows describe $n$ examples and the columns describe $p$ input features as well as $n$-element target vector, $\mathcal{T}$, and a positive integer $k$.

Question: $\exists S \subseteq [1,...,p], |S| \leq k$, such that, $\forall i, j \in [1,...,n]$ where $\mathcal{T}_{i} \neq \mathcal{T}_{j}$, $\exists l\in S$ such that $\mathcal{M}_{i,l}\neq \mathcal{M}_{j,l}$?

That is: is there a cardinality $k$ subset, $S$, of the input features, such that no pair of examples which have identical values across all features in $S$ have a different value for the target feature?

The problem is NP-complete~\citep{davies_np-completeness_1994} and assumed to not be fixed-parameter tractable, when parametrised by $k$, under current complexity assumptions~\citep{cotta_k-feature_2003}. Nonetheless the exact solver used herein solved all instances in significantly less time than was required for subsequently optimising the \gls{fbn}. At least one fast meta-heuristic solver exists for a generalised version of the problem, with reasonable run-times for very large instances~\citep{rocha_de_paula_fast_2015}, and problem instances can be reformulated as instances of the well known Set Cover Problem, for which there are numerous powerful heuristic solvers.

We implemented instances of the \gls{mfs} problem below as Mixed-Integer Programs and solved using them using the Python interface to the IBM CPLEX solver. With a solution procedure in hand we can now define a simple method for estimating the target difficulty order and the level of target overlap.

\subsubsection{\label{sec:ordering_method}Target ordering method}

Given a procedure for solving the \gls{mfs} problem our proposed method for ordering the targets reduces to solving $m$ such instances; one for each target. For all instances the input feature matrix $\mathcal{M}$ is identical to the entire input matrix for the original learning problem. Each instance is solved to optimality resulting in a set of features $S_1, ..., S_m$ for each target.

An estimate of the intrinsic dimension for a target $i$ is then simply $\left|S_i\right|$. Targets are sorted by this estimate with ties broken randomly, and the resulting total order is used in the loss functions defined in Section~\ref{sec:funcs}. These losses could be redefined to treat a subset of targets equally to avoid tie-breaking but we did not explore this possibility.

\subsubsection{\label{sec:nestedness}Target overlap score}

Estimating target overlap also follows from the \gls{mfs} solutions. For two targets, $y_i$ and $y_j$, with solutions, $S_i$ and $S_j$, the target overlap is estimated by the Szymkiewicz-Simpson overlap coefficient:
\begin{equation*}
\sigma\left(i, j\right) = \frac{\left|S_i \cap S_j\right|}{min\left(\left|S_i\right|, \left|S_j\right|\right)} \text{ .}
\end{equation*}
The overlap coefficient gives a value between $0$ and $1$ where $0$ indicates that there is zero overlap between two sets and $1$ indicates that one set is a subset of the other; values in between occur when sets are partially overlapping. This measure was chosen over the more popular Jaccard index as we wish to consider all fully overlapping cases ($A\subseteq B$) equivalent regardless of their relative sizes.

We can then define the \emph{nestedness} score, $\eta$, as the mean overlap coefficient between successive targets (when ordered by the method in Section~\ref{sec:ordering_method}) giving the formula
\begin{equation*}
\eta = \frac{1}{m-1}\sum_{i=2}^{m} \sigma\left(i, i-1\right) \text{ ,}
\end{equation*}
which takes the value $1$ when the \gls{mfs} solutions for all targets forms a perfectly nested sequence of sets, and the value $0$ when the sets have no pairwise overlap. A high value for $\eta$ is thus a necessary but not sufficient condition for strongly inter-dependent Boolean targets. A value near zero indicates that there is unlikely to be any benefit gained from a target curriculum.

The nestedness score allows us to rule out some Boolean \gls{mlc} problems for which a target curriculum would be inappropriate, and the proposed target ordering method gives us a curriculum otherwise. We show experimental results in the following section.

\section{\label{sec:experimental}Experimental Results}

It is important to see how the loss functions defined in Section~\ref{sec:funcs} affect generalisation performance over a range of training set sizes. When optimising $L_1$, generalisation performance for all problems considered transitions from essentially random behaviour to near perfect prediction as training set size increases. With this in mind we generated $1000$ samples each for a range of training set sizes chosen to span that transition by a reasonable margin. We report the training set size as the fraction, $s=\left|\mathcal{I}\right|/2^l$, of all possible patterns.

For each sample we trained one network for each loss function until they achieved zero training error. This allows us to report the difference in test error between the network optimised under $L_1$ and each of the remaining losses, for the same training set. Using the same training set for each loss removes the significant variation in instance difficulty as a confounding factor.

\subsection{\label{sec:problems}Data Sets}

Here we define the problems considered. We first outline problems for which we already have a difficulty curricula. Then we consider several published models of real-world phenomena and finally we test our overall approach on the problem of discovering regulatory network update rules from discretised time-series data.\footnote{Data and code are available at\\\url{github.com/shannonfenn/Multi-Label-Curricula-via-Minimum-Feature-Selection}.}

\subsubsection{\label{sec:problems_circuit}Circuit inference test-beds}

For the first test-bed we chose five circuit-inference problems: binary addition (add) and subtraction (sub) and cascaded variants of three typical circuit-learning and genetic programming problems: multiplexer (cmux), majority (cmaj) and parity (cpar). These were chosen as they all possess a natural hierarchy of targets.

Binary addition and subtraction are well known multi-output problems. For the three remaining problems, multi-output versions are constructed by cascading single-output sub-circuits; each successive output is the result of applying the original function over a larger set of inputs. By this method we construct problems for which we know there is a difficulty order, and what that order is. More detailed definitions of these problems are given in Online Appendix~1.

The cascaded parity and multiplexer possess the strongest dependencies between successive targets, as each subsequent output can be computed directly from the previous output and an additional subset of inputs not already used. The cascaded majority, binary addition and binary subtraction problems have weaker dependencies, requiring extra inputs as well as the previous output and new inputs to compute the next output. This allows us to see the effect of imposing target curricula of problems with varying levels of inter-target dependency.

\Glspl{fbn} are a high variance learner. As such we selected problem sizes so that a meaningful effect size can be seen, but also so that $1000$ instances per loss/sample-size combination could be completed in under a week. Table~\ref{tab:problems} outlines the dimensions selected, as well as the ranges in training sample size.

The hierarchy of targets is known for the above problems since they are constructed from cascading well-known circuits. This ground-truth allows us to further evaluate the methods outlined in Section~\ref{sec:ordering-top} by considering the correlation between this known order and the predicted order.

\subsubsection{\label{sec:problems-real}Boolean Models of Real Phenomena}

In addition to the problems above we also conducted similar experiments using real-world models. This included a commercial integrated circuit: the 74181 4-bit Arithmetic Logic Unit (ALU), and two published biological models.

The 74181 ALU has $14$ inputs and $8$ outputs. It performs $32$ different arithmetic and logical operations on an $8$-bit input with carry in, generating a $4$-bit output with carry out as well as $2$ other carry-related outputs useful for faster calculations when chaining multiple ALUs. The 74181 represents a reverse engineering problem of a realistic size that is less idealised than the above problems, as there is no expected absolute ordering to the targets.

The biological models we considered were Boolean models of the Fanconi Anemia/Breast Cancer (FA/BRCA) pathway and the mammalian cell cycle~\citep{poret_silico_2014}. Boolean regulatory network models like these describe the time-evolution of gene activity, levels of regulatory products and protein activity with a single Boolean variable per element. The system dynamics are then defined with Boolean update functions and an update method (synchronous, partly or fully asynchronous). Both models we consider are synchronous as this enables the inference of a single deterministic update equation by treating successive time steps as input-output pairs. More detail on all models in this section are given in Online Appendix~1.

For these three problems we ran experiments just as in Section~\ref{sec:problems}, with the exception of the ALU problem for which we only trained $100$ networks per training set size due to time constraints. Similarly, due to the large memory requirement of generating all $2^{28}$ patterns for the FA/BRCA model, we used a reduced example pool of $2^{10}$ patterns sampled uniformly. The problem dimensions, training set sizes and example pool sizes are given in Table~\ref{tab:problems}. We have no suspected optimal target order for these problems, so the ordering used for all ordered losses was found using the \gls{mfs}-based intrinsic dimension estimation described in Section~\ref{sec:ordering-top}.

\begin{table}[]
\centering
\begin{tabular}{@{}lcccc@{}}
\toprule
Problem              & Inputs & Targets & Training set size & Example pool size\\  \midrule
Cascaded Parity      & $\ 7$   & $\ 7$   & $[8,\: 112]$   & $\ \ \ 128$ \\
Cascaded Majority    & $\ 9$   & $\ 5$   & $[8,\: 384]$   & $\ \ \ 512$ \\
Binary Subtraction   & $10$    & $\ 5$   & $[8,\: 192]$   & $\ 1,024$ \\
Binary Addition      & $12$    & $\ 6$   & $[8,\: 208]$   & $\ 4,096$ \\
Cascaded Multiplexer & $15$    & $\ 7$   & $[8,\: 180]$   & $32,768$ \\
74181 ALU            & $14$    & $\ 8$   & $[64,\: 384]$  & $16,384$ \\
FA/BRCA pathway      & $28$    &  $28$   & $[8,\: 96]$    & $\ 1,024$ \\
Mammalian cell cycle & $10$    & $\ 9$   & $[8,\: 180]$   & $\ 1,024$ \\
Yeast                & $10$    & $\ 9$   & $ 8 $   & $\ \ \ \ \ \ \ 9$ \\
\emph{E. Coli}      & $10$    & $\ 8$   & $ 4 $   & $\ \ \ \ \ \ \ 5$ \\  \bottomrule
\end{tabular}
\caption{Instance sizes of the test-bed problems.}
\label{tab:problems}
\end{table}

\subsubsection{\label{sec:time-series}Learning Regulatory Networks From Time-Series Data}

While conducting the above experiments we observed that the primary performance gain is seen at smaller sample sizes. For this reason we chose a real-world test-bed where a key issue is the extremely limited amount of available data: inferring Boolean \gls{grn} update models from time-series gene expression data. \cite{barman_novel_2017} provide binarised time-series data for an \emph{E. coli} \gls{grn} and a cell-cycle regulatory network of fission yeast: each consisting of $10$ nodes and only a handful of examples. One desirable result is a synchronous update model such as in the models described in Section~\ref{sec:problems-real}, and one way to achieve this is to treat each sequential pair of states as an input-output pair thus deriving a \gls{fbn} which predicts the next state from the current state. Again, further detail is given in Online Appendix~1.

Some preprocessing of the data were required. First we removed repeat patterns---these occur due to time disparities which we are not attempting to model. Instead we require pairs of distinct patterns as otherwise the requirements for a valid \gls{mfs} instance are not met. Secondly we removed constant targets since they represent nodes for which there is no information suggesting a relationship to any other nodes: one such target was removed from the yeast data, and two from the \emph{E. coli} data.

Finally, since we have very small example sets allowing rapid training, we use leave-one-out cross validation for all patterns whose removal does not introduce additional constant targets (the first was the only such pattern in both cases). For each fold we learned $1000$ networks for each configuration of interest, to account for the significant variance typical in such heavily under-determined problems.

\begin{figure}
    \centering
    \begin{subfigure}[b]{0.34\textwidth}
        \centering
        \includegraphics[height=3.25cm]{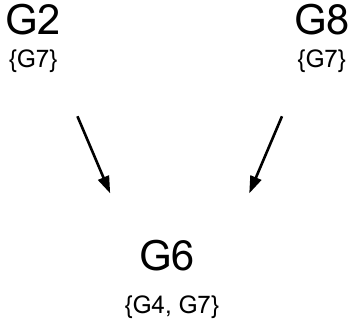}
        \caption{E. Coli}
    \end{subfigure}%
    \hfill
    \begin{subfigure}[b]{0.65\textwidth}
        \centering
        \includegraphics[height=3.25cm]{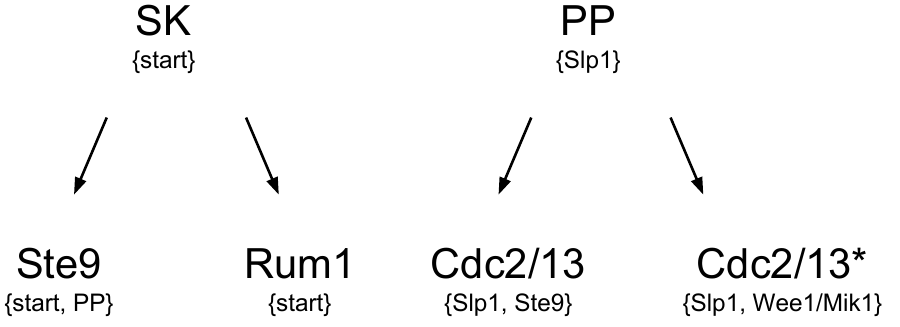}
        \caption{Yeast}
    \end{subfigure}
    \caption{Possible target hierarchies found using \gls{mfs} in the yeast and \emph{E. coli} data sets. In the \emph{E. coli} data (a) we see the target G6 (with feature set $\left\{G4, G7\right\}$) may benefit from being learned after G2 or G8 (both have the feature set $\left\{G7\right\}$). In the yeast data (b) we see two potential hierarchies: either of Ste9 or Rum1 (both with feature set $\left\{start, PP\right\}$) may benefit from being learned after SK, and either of Cdc2/Cdc13 or Cdc2/Cdc13* (with feature sets $\left\{Ste9, Slp1\right\}$ and $\left\{Slp1, Wee1/Mik1\right\}$ respectively) may benefit from first learning PP.}
\label{fig:bio_hierarchies}
\end{figure}

Our initial tests including all targets of the yeast and \emph{E. coli} data sets as a single problem showed little improvement from the ordered losses. Preliminary feature set analysis suggested that there were distinct, shallow hierarchies among the targets (see Figure~\ref{fig:bio_hierarchies}), rather than the deeper hierarchies seen in prior problems. A lack of improvement is unsurprising in this case since imposing an order upon unrelated targets is not expected to offer any benefit and to potentially detriment the performance.

Due to this we opted to learn networks only for the nodes involved in a hierarchy ($2$ for yeast and $1$ for \emph{E. coli}). For fair comparison we compared with the baseline $L_1$ network on both target sets: only those in the suspected hierarchy and all targets. Results are only shown for the former however, since the results of learning a single overall network using $L_1$ were notably worse. Again we used estimated target curricula for all ordered losses (see Section~\ref{sec:ordering-top}).

\subsection{\label{sec:learning}The Learning Procedure}

Here we describe the local search procedure used for training. We learned each function by optimising \glspl{fbn} comprised of only NAND gates using stochastic local search. As the NAND operator is functionally complete the optimisation process does not need to consider node transfer functions and can instead learn any mapping by modifying just the network structure. We set network size, $n_g$, for all problems at $21m$ and represented the structure using a $2\times n_g$ integer matrix, with column $i$ containing the indices of the two sources for the $i$th node.

We ensure the network remains feedforward by imposing a topological ordering of nodes; each node may only accept input from nodes before it in the ordering. Since any \gls{dag} has at least one topological ordering, no structure is excluded by this constraint.

Training involved performing stochastic local search in the space of valid feed-forward structure, using the \gls{lahc} meta-heuristic~\citep{burke_late_2008} with random restarts. This method assumes a black-box view of the learner and is similar in implementation to \gls{sa} while being less sensitive to scaling in the guiding function and requiring only a single meta-parameter in place of a cooling schedule. While the particulars of the optimiser are not of relevance to this work, a brief description is given below and the pseudo-code is given in Online Appendix~1.

\gls{lahc} works similarly to typical stochastic hill climbing: a modification is made to the current solution and accepted based on some criterion. This modification involves a random change to the source of one connection in the \gls{dag}. A modification is accepted if the resulting network has equal or lower loss than the solution obtained $L$ iterations beforehand.

The single meta-parameter, $L$, is the length of a history of costs which provides a dynamic error bound. In preliminary tests, the out-of-sample error was consistent over a wide range in $L$ and so we selected the value which provided the best average convergence speed when using the $L1$-loss. The resulting value was $250$ for all problems, except parity, for which it was $1000$.

For \glspl{mlc} with discrete inputs there are only finitely many possible patterns. Since our test-bed problems are fully defined, we construct each test set from all possible patterns except those used in that particular training set. In doing so the true out-of-sample error for each sample is known.

\subsection{\label{sec:lf_results}Results for Hierarchical Loss Functions}

In this section we present results showing that the losses defined in Section~\ref{sec:funcs} improve the out-of-sample error achieved on all test-bed problems mentioned in Section~\ref{sec:problems}.

\begin{figure}
\centering
\includegraphics[width=0.85\linewidth]{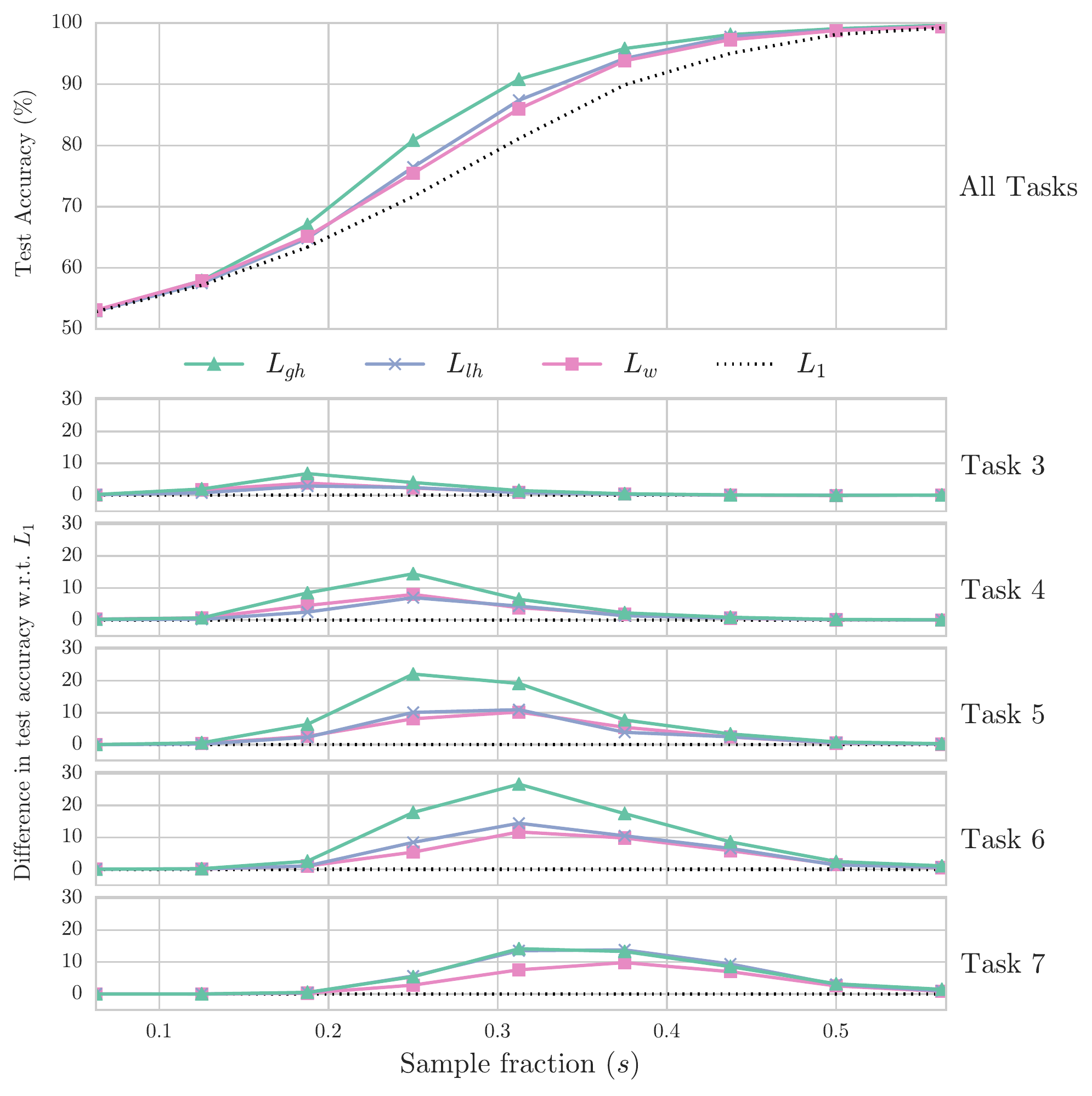}
\caption{Generalisation performance of each loss function on the $7$-bit cascaded parity problem.
The top plot shows the mean test accuracy averaged across all targets for each loss while the remaining plots show the mean difference in test accuracy, on each target, between $L_1$ (dotted baseline) and each of $L_w$, $L_{lh}$ and $L_{gh}$.
The first two targets are not shown as they show only a negligible improvement.
Confidence intervals ($95\%$) are also not shown as they were too narrow to be visible, the widest being less than $2\%$.
A leftward shift in the transition (top) from poor to near-perfect performance can be seen. This becomes more pronounced as the loss more strictly enforces the learning order ($L_w \rightarrow L_{lh} \rightarrow L_{gh}$) with $L_{gh}$ achieving a $16\%$ reduction in the number of training examples needed to achieve above $90\%$ mean test accuracy.
The differences in test accuracy for individual targets (bottom) shows an increase in improvement with target difficulty, as expected; reaching a large peak improvement---from $58\%$ to $84\%$ test accuracy---for the sixth target.
On this particular problem the difference between $L_{lh}$ and $L_{w}$ is not statistically significant.
For all loss functions the out-of-sample accuracy is significantly better than obtained with $L_1$ (dotted line).
}
\label{fig:cpar}
\end{figure}

\begin{figure}
\centering
\includegraphics[width=0.95\linewidth]{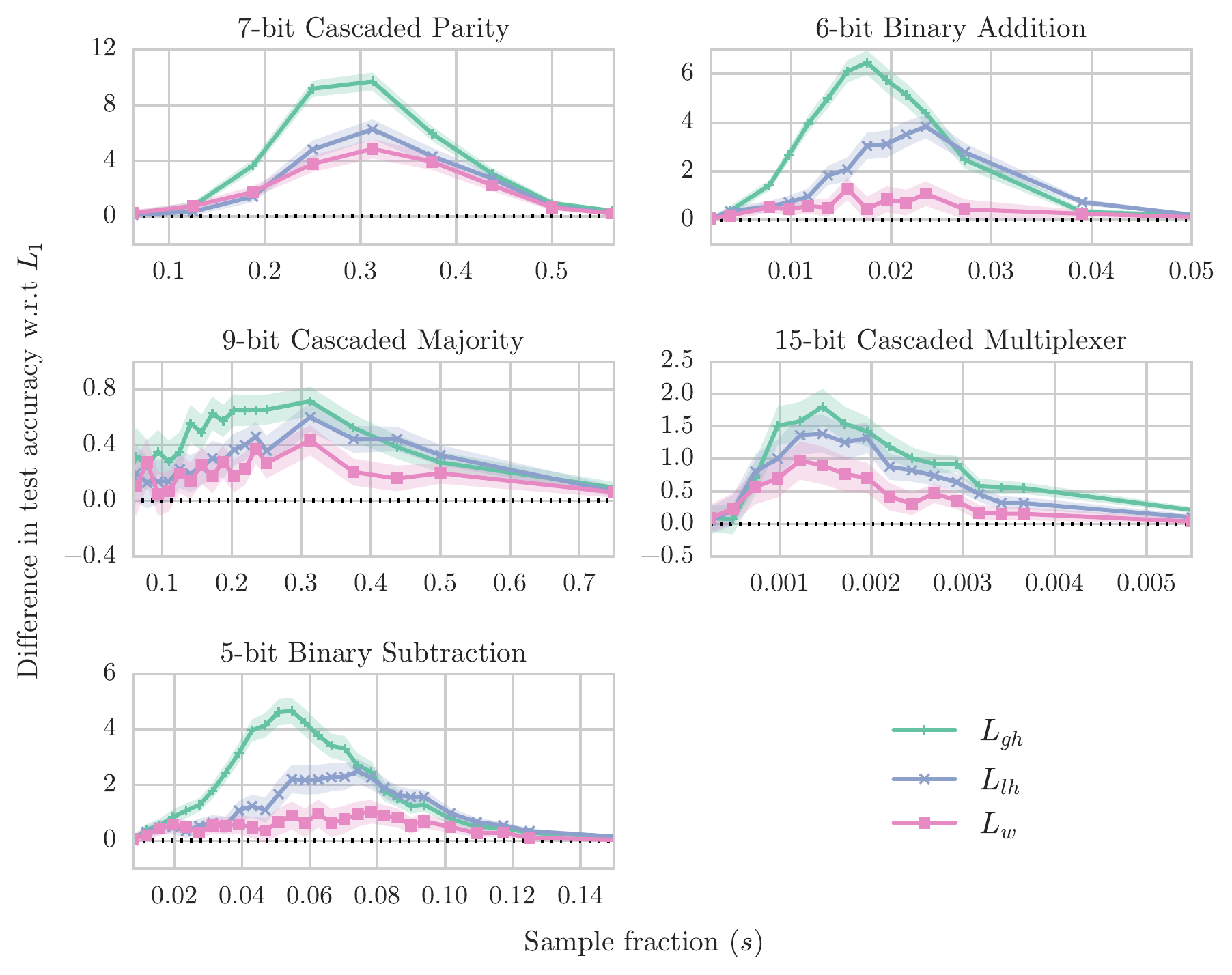}
\caption{Mean difference in test accuracy between $L_1$ (dotted baseline) and each of $L_w$, $L_{lh}$ and $L_{gh}$ on each training sample and problem instance. The $95\%$ confidence interval of the mean is shown with transparent bands. On all problems we can see a statistically significant net increase in test accuracy---though the magnitude varies widely between problems. We can also see that this improvement increases as the loss more strictly enforces the ordering ($L_{w} \rightarrow L_{lh} \rightarrow L_{gh}$).}
\label{fig:diffs}
\end{figure}

Learning \glspl{fbn} with structure-based local search displays high variance. As such, each point shown in Figures~\ref{fig:cpar},~\ref{fig:diffs}, and~\ref{fig:auto} represents the mean performance over $1000$ different training sets of that size. We have also displayed $95\%$ confidence intervals as transparent bands in Figure~\ref{fig:diffs}. For Figure~\ref{fig:cpar} this interval was too small for bars or bands to be visible.

The overall test accuracy as it varies with sample size for the $7$-bit cascaded parity problem is shown in the top of Figure~\ref{fig:cpar}. A phase transition between poor and near perfect generalisation is expected~\citep{patarnello_learning_1987}. What we see in Figure~\ref{fig:cpar} is a leftward shift in this transition as we impose an increasingly strict order to the targets. Less examples are required to achieve the same generalisation if a difficulty-based target curriculum is imposed.

Even for a single problem instance the improvement is not expected to be consistent across all targets. In fact our earlier reasoning suggests the improvement should increase for successive targets. In the bottom of Figure~\ref{fig:cpar} we have also shown the test accuracy improvement given by each loss, for each target of $7$-bit cascaded parity. The first two targets are not shown; as expected there is little improvement for them. Beyond this, we see that the improvement increases with each successive target until the last (the same trend appears in all five problems). This peaks at the second-last target with an increase from $58\%$ test accuracy using $L_1$, to $84\%$ test accuracy using $L_{gh}$.

Generalisation improvements can be most readily observed when looking at the test accuracy differences between $L_1$ and each of $L_w$, $L_{lh}$ and $L_{gh}$ as they vary with $s$. Figure~\ref{fig:diffs} shows the mean of these differences with respect to $s$ along with transparent bands indicating the $95\%$ confidence interval of the mean. We can see that the improvement differs significantly between targets but is consistently positive and statistically significant across a range of training set sizes for all five problem instances.

More importantly, the loss which most strictly enforces the learning order:~$L_{gh}$, also confers the largest improvement, and vice versa for the loss which least strictly enforces it:~$L_{w}$. This phenomena is consistent across all five problems and confirms the central idea presented in this paper.

In terms of overall generalisation accuracy we can see that training to a curriculum of targets offers a significant advantage in Boolean \glspl{mlc} with target hierarchies. Furthermore we can see that a curriculum can be enforced in a combinatorial black-box optimiser using only slight modifications to the typical guiding function.

\subsection{\label{sec:eval_order}Evaluating the Discovered Curricula}
Here we evaluate the effectiveness of the proposed methods for estimating target overlap and difficulty order described in Section~\ref{sec:order} on the same test-bed problems. We also observe how the improvement conferred by hierarchical loss functions varies when random orderings are used.

\subsubsection{\label{sec:tau1}Effectiveness of order estimation}

We know the difficulty curricula for the problems in Section~\ref{sec:problems_circuit}. Thus we can quantify the effectiveness of the proposed target ordering method on these, by using the rank correlation between the known and estimated orderings. Since these are permutations there are no ties to consider and we can use the simplest version of Kendall's $\tau$:
\begin{equation*}
\tau = \frac{P - Q}{P + Q} \text{ ,}
\end{equation*}
where $P$ is the number of target pairs which are placed in the same relative order under both orderings, and $Q$ the number which are not. Being a correlation, $\tau$ takes values in $[-1, 1]$, with $1$ indicating the two orderings are identical, $0$ indicating they are uncorrelated and $-1$ that they are inverses.

\begin{figure}
\centering
\includegraphics[width=0.95\linewidth]{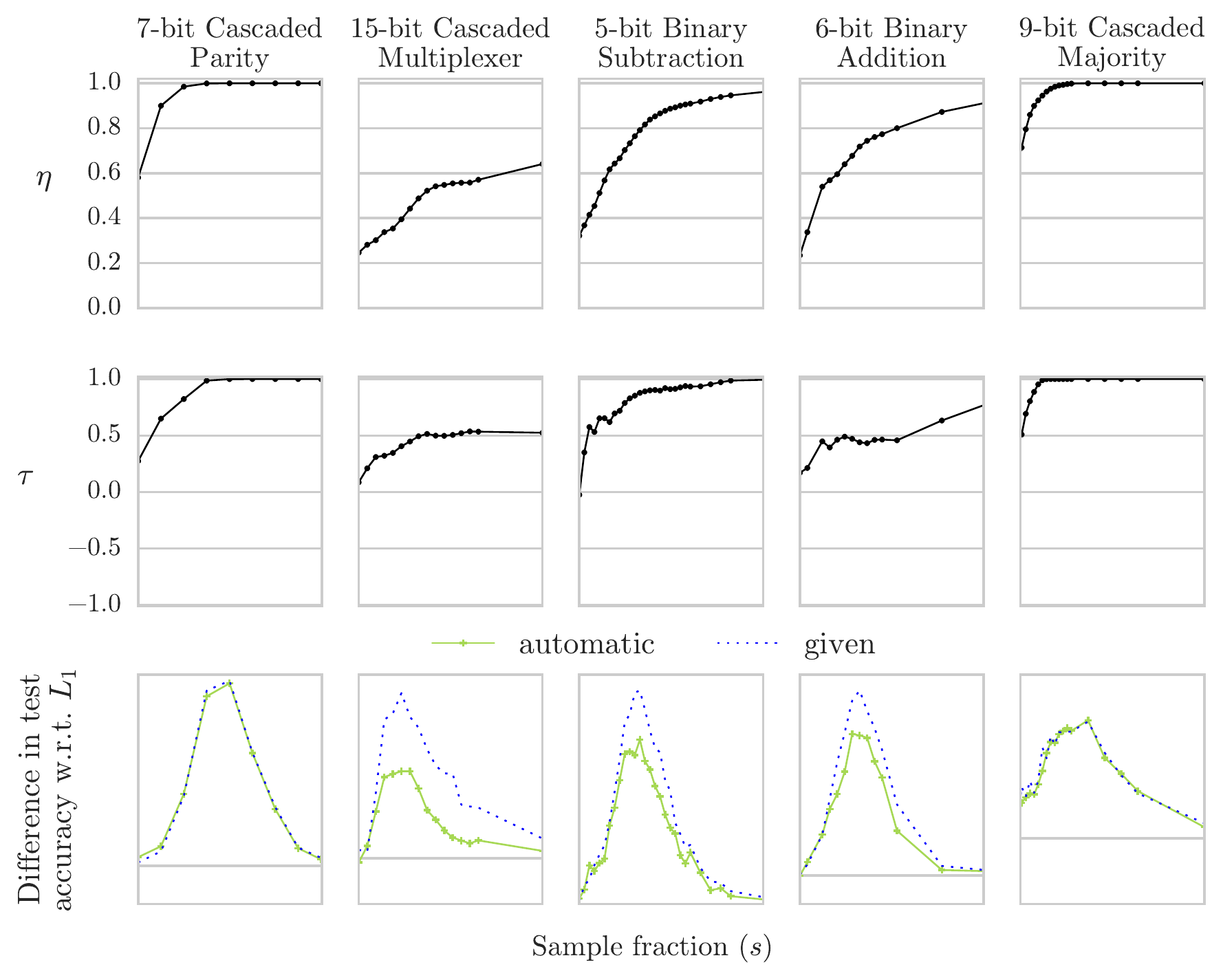}
\caption{Mean increase in test accuracy---over networks trained using $L_1$---for networks trained using $L_{gh}$ with the known target order (given) and the automatically detected target order (automatic).
For all problems except CMUX both $\tau$ and $\eta$ reach values above $0.5$ before the respective peaks in test error difference.
For these same problems the detected target order provides the nearly the same benefit as the known target order, which is no surprise given the strong correlation between the known and estimated orderings. The problem with the largest deviation, CMUX, is also the one for which $\tau$ and $\eta$ are weakest.
Note: the x-axes are only shared within columns, and the y-axis for the third row is not shared, since some problems have a much larger improvement than others.}
\label{fig:auto}
\end{figure}

We used the same experimental configuration as before but for each training instance we randomly shuffled the targets, to reduce the effect of any deterministic biases. Then we estimated an ordering using the \gls{mfs}-based method and computed the rank correlation between this ordering and the order used in Section~\ref{sec:lf_results}, as well as the nestedness score, $\eta$, defined in Section~\ref{sec:nestedness}. Finally we learned one network for each instance using $L_{gh}$ with the estimated target order. Only $L_{gh}$ was considered since it produced the largest improvement with the known order (see Figure~\ref{fig:diffs}). These results are shown in Figure~\ref{fig:auto}.

For each problem, Figure~\ref{fig:auto} shows the nestedness score and the rank correlation between the known and automatically discovered target orders; as well as the generalisation improvement imparted by $L_{gh}$ when given each ordering.

We see that not all problems are easily detected as possessing overlapping hierarchical targets. On parity and majority the mean nestedness score and $\tau$ quickly reach $1.0$ indicating that the proposed method is successful at automatically determining that there is an overlap, and the correct order of targets. For addition and subtraction the results are weaker but still promising however for the multiplexer there is only a weak correlation between the expected and known target orders and a low nestedness score. It is interesting that, even with a weakly correlated target order, some improvement is still gained from enforcing that order.

Except for CMUX---for which there is still an overall improvement---the third row in Figure~\ref{fig:auto} shows a high agreement between test accuracies gains using given and automatically detected target orderings. With zero prior knowledge of the respective target difficulties, the \gls{mfs}-based method combined only with a slight change in the loss function, yields significant improvements in generalisation. This is promising given we are considering a learner which was already leveraging the expected benefits associated with a shared internal representation.

\subsubsection{\label{sec:tau2}How performance correlates with target order}

The next experiment we describe had two principal aims: to examine the relationship between the performance of hierarchical loss functions and the given curriculum, and to determine if the difficulty based curriculum that we have assumed as ground truth is actually optimal.

To do this we used a slightly smaller single problem---$5$-bit addition---and trained networks using $L_{gh}$ with randomly generated target permutations and $L_1$. This gives us another important baseline---random target orderings---and also allows us to see how the performance of $L_{gh}$ varies with $\tau$, and if the least-to-most-significant curriculum we propose is outperformed by some other ordering.

We did not draw permutations uniformly since the rank correlation, $\tau$, is heavily skewed over the space of permutations; the values $\pm 1$ are only represented by $2$, in combinatorially many, possibilities. However for a $n$-target permutation there are only $\left\lceil{n\times(n-1)/2}\right\rceil$ unique values of $\tau$, so selecting permutations uniformly distributed over $\tau$ is trivial. As such we trained $5$ networks for each of $50$ permutations (selected with replacement) for every value of $\tau$ over the range of sample sizes. The baseline $L_1$ does not vary with order so we again learned $1000$ networks per sample size.

\begin{figure}
\centering
\includegraphics[width=0.95\linewidth]{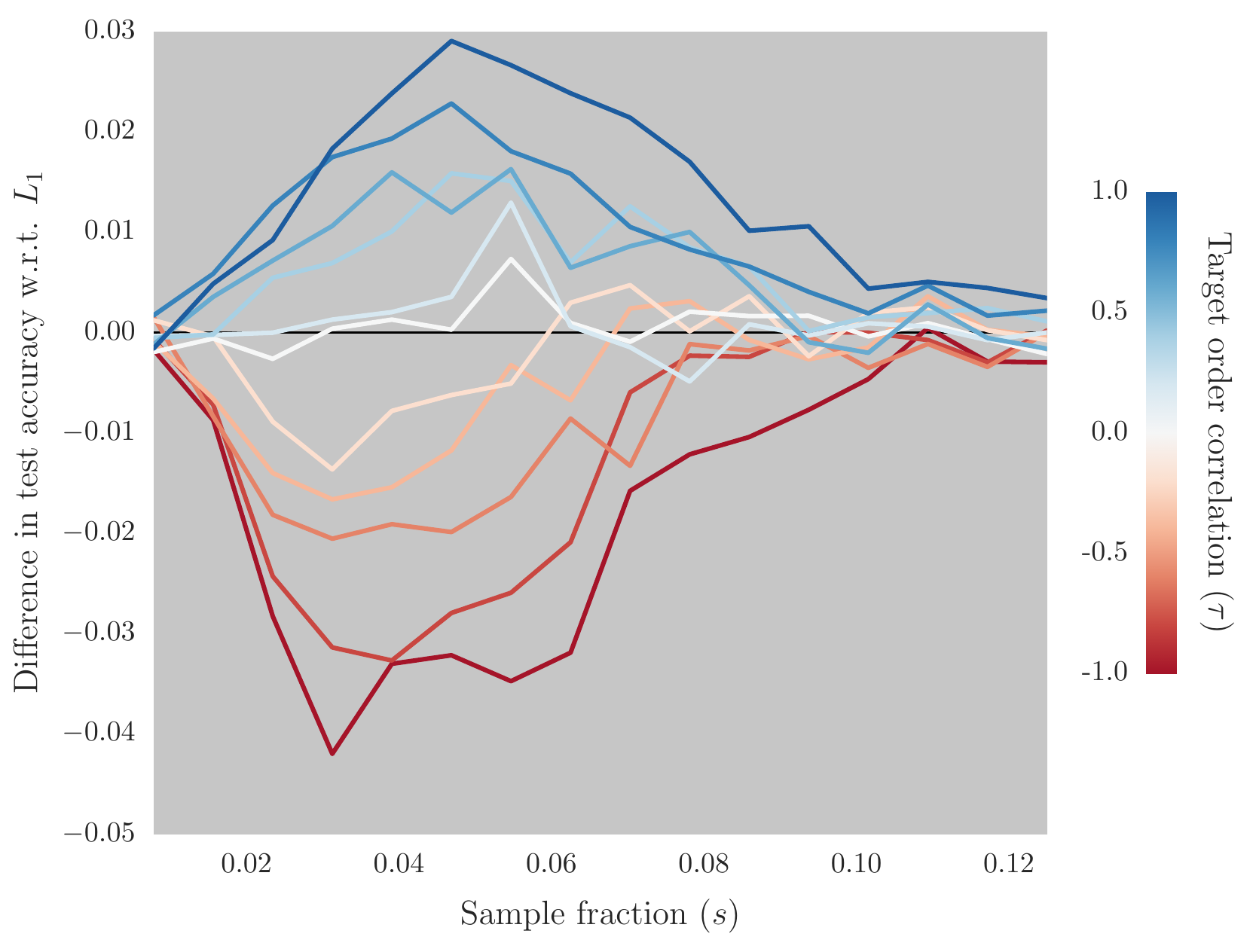}
\caption{Mean difference in test accuracy between $L_{gh}$ and $L_1$ over various training set sizes for the $5$-bit addition problem for randomised target orderings. Each line represents the mean change in test accuracy for each possible level of correlation between the randomly given order and the known order. For an $n$-element permutation there are $\left\lceil{n\times(n-1)/2}\right\rceil$ possible values of $\tau$, hence $11$ lines. The central permutations offer little benefit, and since the random permutations are skewed heavily toward $\tau = 0$ we can infer that a randomly generated ordering would offer no improvement on average. We see very clearly that that direct opposite ($\tau = -1$) of the suggested curricula impeded the learning performance by as much as the correct order improved it and that the exact easy-to-hard ordering ($\tau = 1$) conferred the largest benefit.}
\label{fig:err-wrt-tau}
\end{figure}

Figure~\ref{fig:err-wrt-tau} shows the resulting difference in test accuracy. Each line shows the mean improvement (over networks trained using $L_1$) as it varies with sample size for networks trained with $L_{gh}$ for all permutations with that correlation value. One clear observation is that permutations which positively correlate (blue) with the easy-to-hard ordering give improvement while permutations which negatively correlate (red) actually detriment performance. The second key observation is that this varies with the strength of the correlation; and, most importantly, the largest improvement is given by the exact easy-to-hard curriculum ($\tau = +1$).

These results confirm our intuition that an easy-to-hard ordering is optimal---at least for this method of enforcing a target curriculum. They also agree with the results seen in Section~\ref{sec:tau1} which showed that some improvement was seen for the CMUX problem even with a weak mean order correlation, but also that weaker correlation in the discovered curriculum reduced the improvement given by $L_{gh}$.

\subsection{\label{sec:real_results}Real-World Problems}

Here we present results for the experiments described in Sections~\ref{sec:problems-real} and~\ref{sec:time-series}. We lack an expected target order so the results for all ordered losses are with automatically detected curriculum (Section~\ref{sec:ordering-top}).

\begin{figure}
\centering
\includegraphics[width=0.95\linewidth]{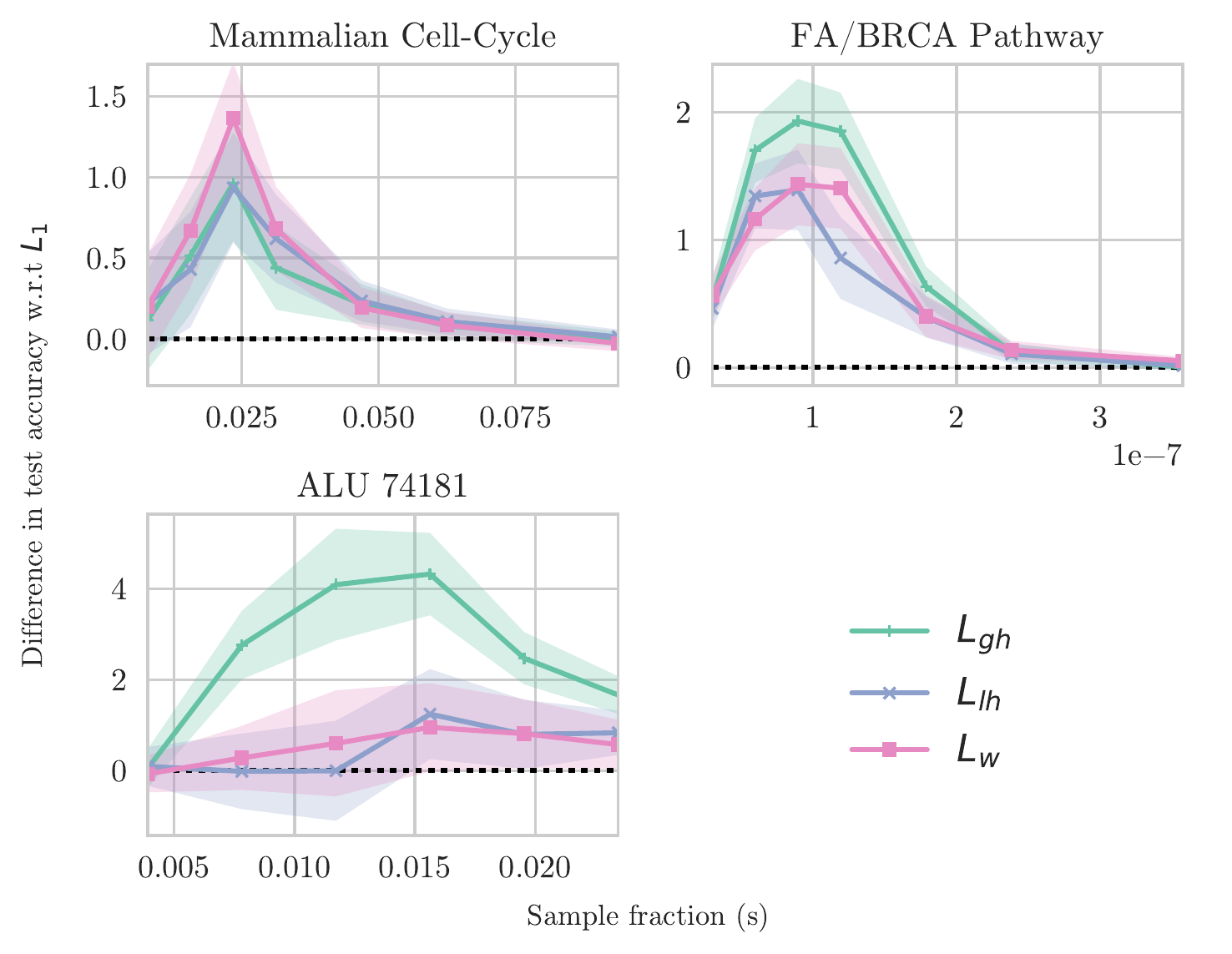}
\caption{Mean difference in test accuracy between $L_1$ (dotted baseline) and each of $L_w$, $L_{lh}$ and $L_{gh}$ on each training sample for the 74181, mammalian cell-cycle and FA/BRCA pathway models. The $95\%$ confidence interval of the mean is shown with transparent bands. On all problems we see the same net increase in test accuracy---again varying widely between problems. We see again that, except in the case of the mammalian cell-cycle model for which the losses are not clearly distinguishable, this improvement increases as the loss more strictly enforces the ordering ($L_{w} \rightarrow L_{lh} \rightarrow L_{gh}$).}
\label{fig:diffs-amf}
\end{figure}

For the ALU, mammalian cell-cycle, and FA/BRCA pathway models we have the same regime as for the prior test-beds, so we have presented the results in the same manner. Figure~\ref{fig:diffs-amf} shows the mean difference in test accuracy between each hierarchical loss function and the baseline $L_1$ as sample size is varied. We also see that with the mammalian cell-cycle model there is much less differentiation between the hierarchical losses, in fact $L_{w}$ appears to have given the most benefit. Overall, in all three cases imposing the target order improved test accuracy and the proposed method clearly discovered a beneficial curriculum.

For the time-series data sets the leave-one-out testing regime did not involve varying the sample size so we have instead presented the results in Tables~\ref{tab:yeast} and~\ref{tab:ecoli}. Again we see that imposing a target curriculum improved the generalisation performance of some targets later in the suspected order even for shallow hierarchies and with extremely limited sample sizes. In Table~\ref{tab:ecoli} we see an improvement in the target G6 and in Table~\ref{tab:yeast} we see improvement for both subsequent targets---Cdc2/Cdc13 and Cdc2/Cdc13*---in one of the two estimated hierarchies (see Figure~\ref{fig:bio_hierarchies}).

\begin{table}[]
\centering

\begin{tabular}{cccccccc}
\toprule
     &   SK     &   Ste9    &   Rum1  &    PP     & Cdc2/Cdc13 & Cdc2/Cdc13* \\ \midrule
$L_{w}$   &   0.4    &    0.5    &    0.4  &    1.6    &    1.2     &    0.6\\
$L_{lh}$   &   0.6    &    0.5    &    0.4  &    0.8    &    1.0     &    0.1\\
$L_{gh}$   &   0.9    &    0.5    &    0.3  &    0.5    &    1.3     &    1.4 \\ \bottomrule
\end{tabular}
\caption{Results for the yeast data set. See Figure~\ref{fig:bio_hierarchies} for the estimated hierarchies. The SK$\rightarrow$ \{Ste9, Rum1\} hierarchy has given little improvement, however we see more promise in the PP$\rightarrow$ \{Cdc2/Cdc13, Cdc2/Cdc13*\} hierarchy, particularly for $L_{gh}$.}\label{tab:yeast}

\bigskip

\centering
% \begin{tabular}{@{}cccc@{}}
\begin{tabular}{cccc}
\toprule
     &   G2     &    G8     &    G6 \\ \midrule
$L_{w}$   &   0.1    &    0.2    &    0.8 \\
$L_{lh}$   &   0.7    &   -0.9    &    1.3 \\
$L_{gh}$   &   0.1    &    0.1    &    1.4 \\ \bottomrule
\end{tabular}
\caption{Results for the \emph{E. coli} data set. See Figure~\ref{fig:bio_hierarchies} for the estimated hierarchies. The \{G2, G8\}$\rightarrow$G6 hierarchy has given some improvement, however for $L_{lh}$ we see a drop in performance on one of the base targets in the hierarchy. For $L_{gh}$ the results are still overall positive.}
\label{tab:ecoli}
\end{table}

\section{\label{sec:discussion}Discussion and Future Work}

We have shown that loss functions that impose a target order improve the overall generalisation performance of \glspl{fbn} using a series of hierarchical multi-target circuit-inference problems. We also showed this improvement increases as the target order is more strictly enforced. Note that the losses considered do not alter the set of global optima. This approach does not restrict the space of networks but instead modifies their probability of discovery in a natural way using concepts inspired by human teaching and learning practices.

The set of circuit-inference problems described in Section~\ref{sec:problems_circuit} all possess a common element: intermediate values which are useful for computing the $m+1^{th}$ output are computed for the $m^{th}$ output. A human designer with access to a complete $m$-bit circuit would have a much easier time building an ($m+1$)-bit circuit. It is natural to assume that this ease should extend---in terms of convergence speed, generalisation or both---to a optimiser which builds such circuits in the correct order. We observed the latter but not the former.

We observed that performance improvements primarily occur around the existent generalisation phase transition. This is intuitive: there are fundamental limits with respect to training-set size that cannot be alleviated, and there are also sizes for which there is enough information present that teaching a target order confers no advantage. What we have shown is that---in the interim region between too little and plenty of training information---teaching targets in order of difficulty can make better use of what data is present. This improvement is expected in light of the positive effects of hierarchical cost functions on optimiser performance~\citep{moscato_introduction_1993-1}.

The improvement on the final target from $L_{gh}$ actually decreases (bottom of Figure~\ref{fig:cpar}). Initially, this appears as a cause for concern as it seems to indicate a point of diminishing returns. However this happened consistently only on the final target for all problems. Were there a per-problem tipping point we would expect to see it before the last target, or not at all, on some problems. This is also the same for varying sizes of binary addition (see Online Appendix~2).

A more likely explanation is the gate budget coupled with a lack of constraints on which gates each target can source from. The loss function for which we see the issue occurring, $L_{gh}$ essentially forces optimisation of each target in succession. As the process continues, it becomes increasingly more likely that any given target will have a majority of its computational nodes closer to the end of the ordering of gates. Subsequent sub-networks will thus require later and later nodes to be available in order to correctly discover the hierarchical structure of the problem which becomes an issue toward the end when the gate budget is exhausted. This could be alleviated by reserving nodes for each target.

We have also suggested an intuitive approach to determining the target overlap for Boolean \glspl{mlc}, and to automatically determine an appropriate target order. For the problems considered, the bulk of the generalisation improvement conferred by the curriculum remained, even when no information on the true target order was provided to the learner.

We further examined the same loss functions in conjunction with the target curricula estimation method on three models of real world phenomena as well as on two regulatory network time-series data sets. On all three models we observed test accuracy increases similar to that seen on the circuit-inference problems. In the case of the yeast and \emph{E. coli} data set we discovered possible target hierarchies which were shallow and as such we expected less improvement (Figure~\ref{fig:cpar} shows that generalisation gains are focused primarily on later targets). Nonetheless the results are still in line with what we have observed for the previous problems. This is promising as we have not explicitly accounted for noise in either the curricula detection method or design of the loss functions.

Finally we examine the importance of the curricula itself in Section~\ref{sec:tau2}. We see that the easy-to-hard curricula gave the strongest improvement, that curricula which were anti-correlated with the easy-to-hard permutation actually decreased test accuracy, and that the curricula which were uncorrelated (i.e. essentially random) gave no improvement over learning with no curriculum. The fact that the peak in the distribution of permutations over $\tau$ occurs at zero means that randomly generated curricula will most commonly give no benefit and thus highlights the importance of principled methods for determining target curricula.

Not discussed here is the effect that the use of these guiding functions has on the average convergence time. We observed a variable effect on training speed and, for some problems, there is a cost in convergence speed when using target curricula. More information on training time is available in Online Appendix~2.

Another possible impact on training speed is the requirement for the target ordering method to solve an NP-hard optimisation problem. However, for all problems considered, the target order detection process had almost no impact on overall training time. For significantly larger problem instances, however, heuristic solvers would likely be required.

Noise has not been considered as a factor in this work. It is uncertain whether the \gls{mfs} based methods for estimating target overlap and target schedules---or \glspl{fbn} themselves---are robust to incorrect target labels. Future work should address this shortfall, perhaps by allowing the per-target error constraints in $L_{gh}$ to be non-zero. The drop in performance on the final target should also be addressed; possibly by using stratified networks, or annealed depth constraints, to prevent later gates being consumed by earlier targets.

While our results are only for the case of Boolean \glspl{mlc} learned with \glspl{fbn}, we expect they may extend to other non-convex learners, such as deep neural nets, for problems with categorical or Boolean domains, provided differentiable analogues to $L_{lh}$ and $L_{gh}$ are found. Future work could also examine target feature selection methods for real valued domains, as well as heuristic Boolean feature selection methods, to determine if the approach presented here is more generally applicable. A recent discussion of the use of intrinsic dimension as a proxy for the overall complexity of arbitrary data sets~\citep{granata_accurate_2016} is promising.

\acks{We thank Ms Natalie de Vries for her assistance with proof reading and discussion of the manuscript, and Drs. Alexandre Mendes and Nasimul Noman for their fruitful discussions in the early phases of the work. Pablo Moscato also wishes to thank Prof. Miguel \'{A}ngel Virasoro for a preprint copy of \cite{patarnello_learning_1987} many years ago. P.M. acknowledges funding of this research by the Australian Research Council grants Future Fellowship FT120100060 and Discovery Project DP140104183 and the Maitland Cancer Appeal.}

\vskip 0.2in
\bibliography{17-007}

\end{document}

\title{Target Curricula via Selection of Minimum Feature Sets:\\a Case Study in Boolean Networks (Online Appendix 1)}

\author{\name Shannon Fenn \email shannon.fenn@newcastle.edu.au \AND
       \name Pablo Moscato \email pablo.moscato@newcastle.edu.au \\
       \addr School of Electrical Engineering and Computing\\
       University of Newcastle \\
       University Drive \\ Callaghan NSW 2308 \\ Australia}

\editor{Amos Storkey}

\maketitle

\section{Test-bed problems}

Here we give more detailed definitions of the test-bed problems discussed in the main text. In the following definitions $\wedge$ denotes conjunction, $\vee$ inclusive disjunction, $\oplus$ exclusive disjunction and $\overline{x}$ negation.

\subsection{Problems with known target curricula}

Binary addition takes two $n$-bit inputs, $x$ and $y$, and computes an $n$-bit\footnote{We ignore initial carry in and final carry out} output $z$. The $i^{th}$ output can be expressed as $z_i = x_i \oplus y_i \oplus c_{i-1}$ where $c_i$ represents an intermediate \emph{carry} value given by $c_i = \left(x_i \wedge y_i\right) \vee \left(c_{i-1} \wedge \left(x_i \oplus y_i\right)\right)$. An $n$-bit addition results in a problem with $N_i = 2n$ and $N_o = n$.

Binary subtraction is similarly defined. The $i^{th}$ output can be expressed by $z_i = x_i \oplus y_i \oplus b_{i-1}$ which is almost identical to addition but with $b_i$ representing an intermediate \emph{borrow} value given by $b_i = \left(\overline{x_i} \wedge y_i\right) \vee \left(b_{i-1} \wedge \overline{\left(x_i \oplus y_i\right)}\right)$.

The three remaining problems are built by cascading well-known single output circuits. Each successive output is the result of applying the original function over a larger set of inputs. By this method we build circuits of different flavours for which we know there is a difficulty order, and what that order is.

The multiplexer is a generalised switch which takes $n$ data inputs, $d$, and $\lceil log_2\left(n\right)\rceil$ select inputs, $s$, with output equal to the value of the $i_{th}$ data input where $i$ is the integer value given by the select inputs. The output ($z$) of the cascaded variant is defined by
\begin{align*}
z_0 &= \left(d_0 \wedge \overline{s_0}\right) \vee \left(d_1 \wedge s_0\right) \text{ and}\\
z_i &= \left(z_{i-1} \wedge \overline{s_i}\right) \vee \left(d_i \wedge s_i\right) \text{.}
\end{align*}
For any given $n$ the resulting problem has dimensions $N_i = 2n-1$ and $N_o = n-1$.

The majority function takes an $n$-bit input, $x$, and gives a single output which takes the value of the majority of the input bits. The output ($z$) of the cascaded variant is given by
\begin{equation*}
z_i = \begin{cases}
  0, & \text{if $\frac{i}{2} < \sum_{j=0}^{i+1} x_j$} \\
  1, & \text{otherwise}
\end{cases}\text{ ,}
\end{equation*}
resulting in a problem with $N_i = n$ and $N_o = \lceil \frac{n}{2}\rceil$.

The final problem, parity, is likely the most well known. It takes the value $1$ when the number of $1$s in the input is \emph{odd} and $0$ otherwise. The cascaded version presents $N_i$ outputs given by $z_i = \left(\sum_{j=0}^{i} x_j\right) mod\:2$.

\subsection{Problems with unknown target curricula}
\subsubsection{74181 ALU}
This is a model of an $8$-bit Arithmetic/Logic Unit. It has 14 input lines and 8 output lines. It performs $32$ different arithmetic and logical operations on an $8$-bit input with carry in, generating a $4$-bit output with carry out as well as $2$ other carry-related outputs useful for faster calculations when chaining multiple ALUs. Datasheets for this IC are publicly available.

\subsubsection{Mammalian cell-cycle model}
The full model is given in the references cited in the main text. It consists of $10$ nodes, and thus $10$ inputs, and $10$ targets. However in the published model the Cyclin D node does not update and is treated as a constant input, as such we do not treat it as a learnable target.

\subsubsection{FA/BRCA pathway model}
The full model is given in the references cited in the main text. It is much larger consisting of $28$ nodes, and thus $28$ input and $28$ target features.

\subsection{Time-series Regulatory Network data}
The yeast dataset consists binarized time-step values for $10$ nodes: start, SK, Cdc2/Cdc13, Ste9, Rum1, Slp1, Cdc2/Cdc13*, Week1Mik1, Cdc25, PP, and Phase. The ``start'' target was constant and thus removed.

The E. coli dataset also consisted binarized time-step values for $10$ nodes G1 up to G10. We found the targets G7 and G10 were constant and thus removed them.

\section{Late-Acceptance Hill Climbing variant}

In this section we provide pseudo-code for the Late-Acceptance Hill Climbing variant used in this work. The algorithm has a single meta-parameter, $L$, representing the length of a cost-history list. The value of $L$ used for each problem instance is given in the main text.

\begin{algorithm}[H]
 \caption{LAHC implementation}
 \label{alg:lahc}
    \begin{algorithmic}[1]
    \Require
        \Statex A scalar cost function: $C()$
        \Statex An initialisation method: $initialise()$
        \Statex A neighbour generation method: $neighbour()$
        \Statex A cost history length: $L>0$
        \Statex An iteration limit: $I>0$
        \Statex A restart limit: $R\geq 0$
    \Statex
    \State $r\gets 0$
    \Repeat
      \State $s\gets initialise()$
      \ForAll{$k\in \left\{0, \dots, L-1\right\}$}
          \State $\hat{C}_k\gets C\left(s\right)$
      \EndFor
      \State $i\gets 0$
      \Repeat
          \State $s^*\gets neighbour(s)$
          \State $v\gets i \bmod L$
          \If {$C\left(s^*\right) < \hat{C}_v$ or $C\left(s^*\right) \leq C\left(s\right)$}
              \State $s\gets s^*$
          \EndIf
          \State $\hat{C}_v\gets C\left(s\right)$
          \State $i\gets i + 1$
      \Until{$i = I$ or $C\left(s\right) = 0$}
    \Until{$r = R$ or $C\left(s\right) = 0$}
    \end{algorithmic}
\end{algorithm}

\title{Target Curricula via Selection of Minimum Feature Sets:\\a Case Study in Boolean Networks (Online Appendix 2)}

\author{\name Shannon Fenn \email shannon.fenn@newcastle.edu.au \AND
       \name Pablo Moscato \email pablo.moscato@newcastle.edu.au \\
       \addr School of Electrical Engineering and Computing\\
       University of Newcastle \\
       University Drive \\ Callaghan NSW 2308 \\ Australia}

\editor{Amos Storkey}

\maketitle

\begin{figure}[h]
\centering
\includegraphics[width=0.85\linewidth]{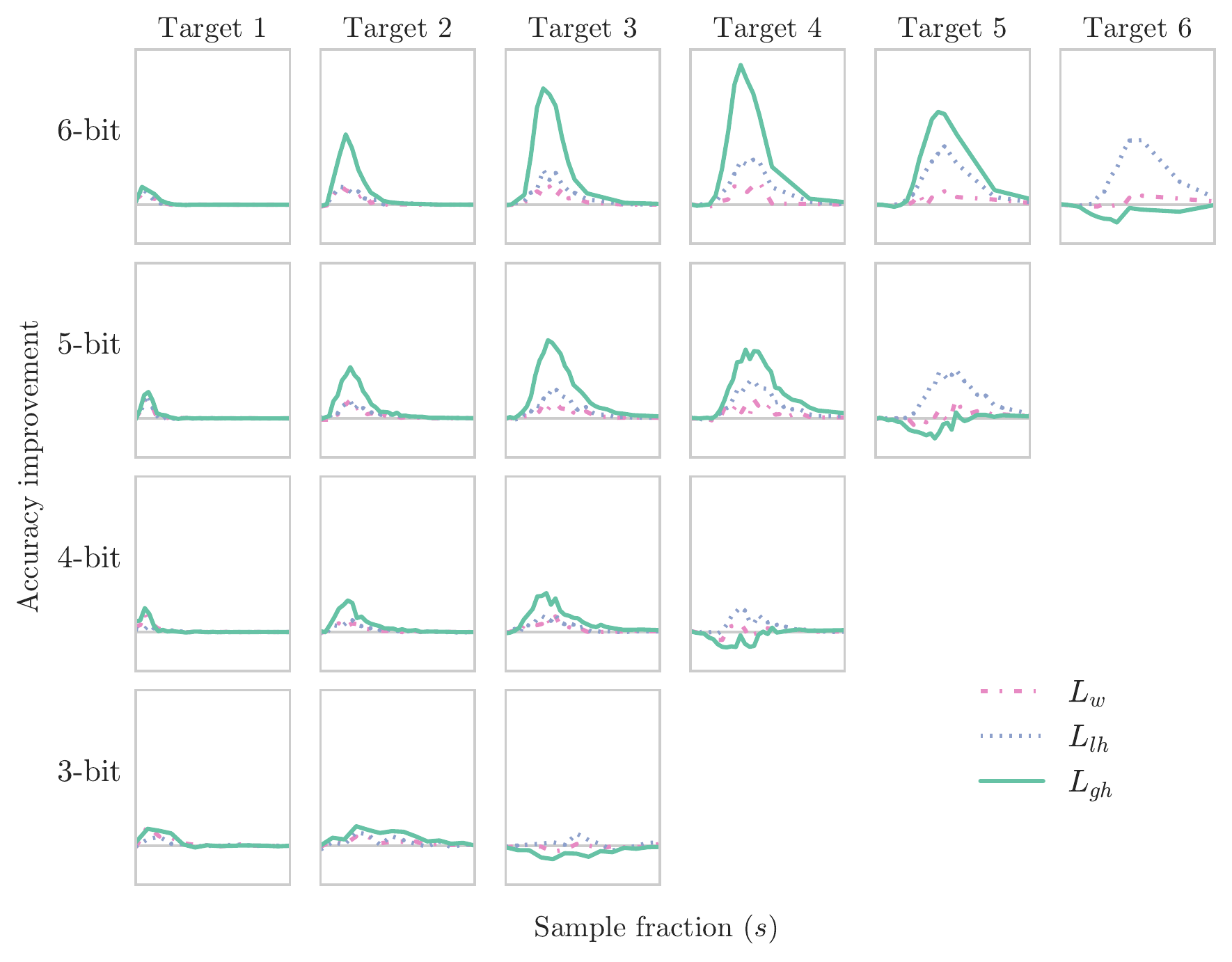}
\caption{Mean increase in test accuracy---over networks trained using $L_1$---for networks trained using $L_{w}$, $L_{lh}$, and $L_{gh}$ on the $3$-, $4$-, $5$-, and $6$-bit binary addition problem. The y-axes all possess the same range, and the x-axis are shared along each row.}
\label{fig:add}
\end{figure}

\begin{figure}
\centering
    \begin{subfigure}[t]{0.5\textwidth}
        \centering
        \includegraphics[width=0.9\linewidth]{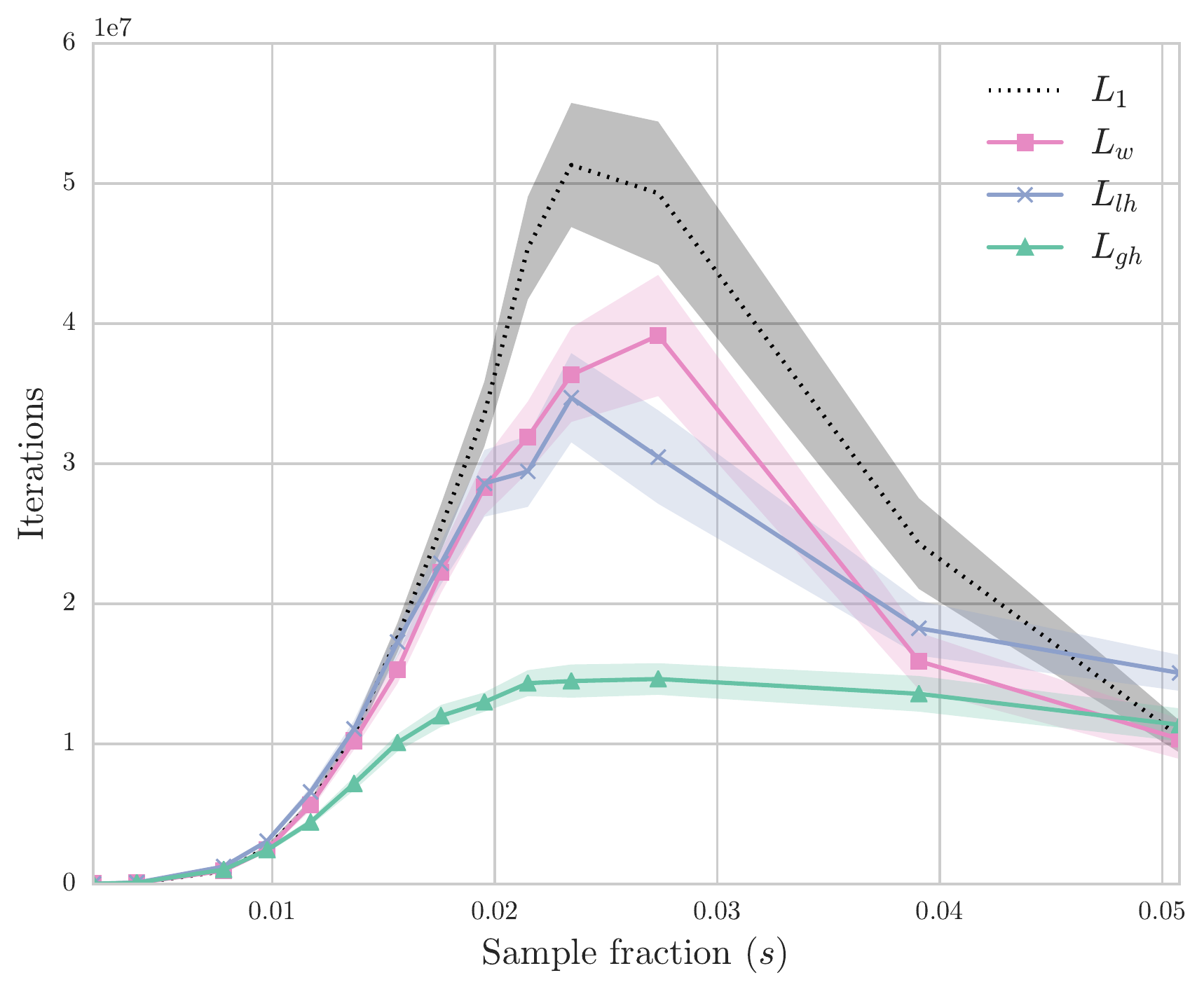}
        \caption{$6$-bit binary addition}
    \end{subfigure}%
    ~
    \begin{subfigure}[t]{0.5\textwidth}
        \centering
        \includegraphics[width=0.95\linewidth]{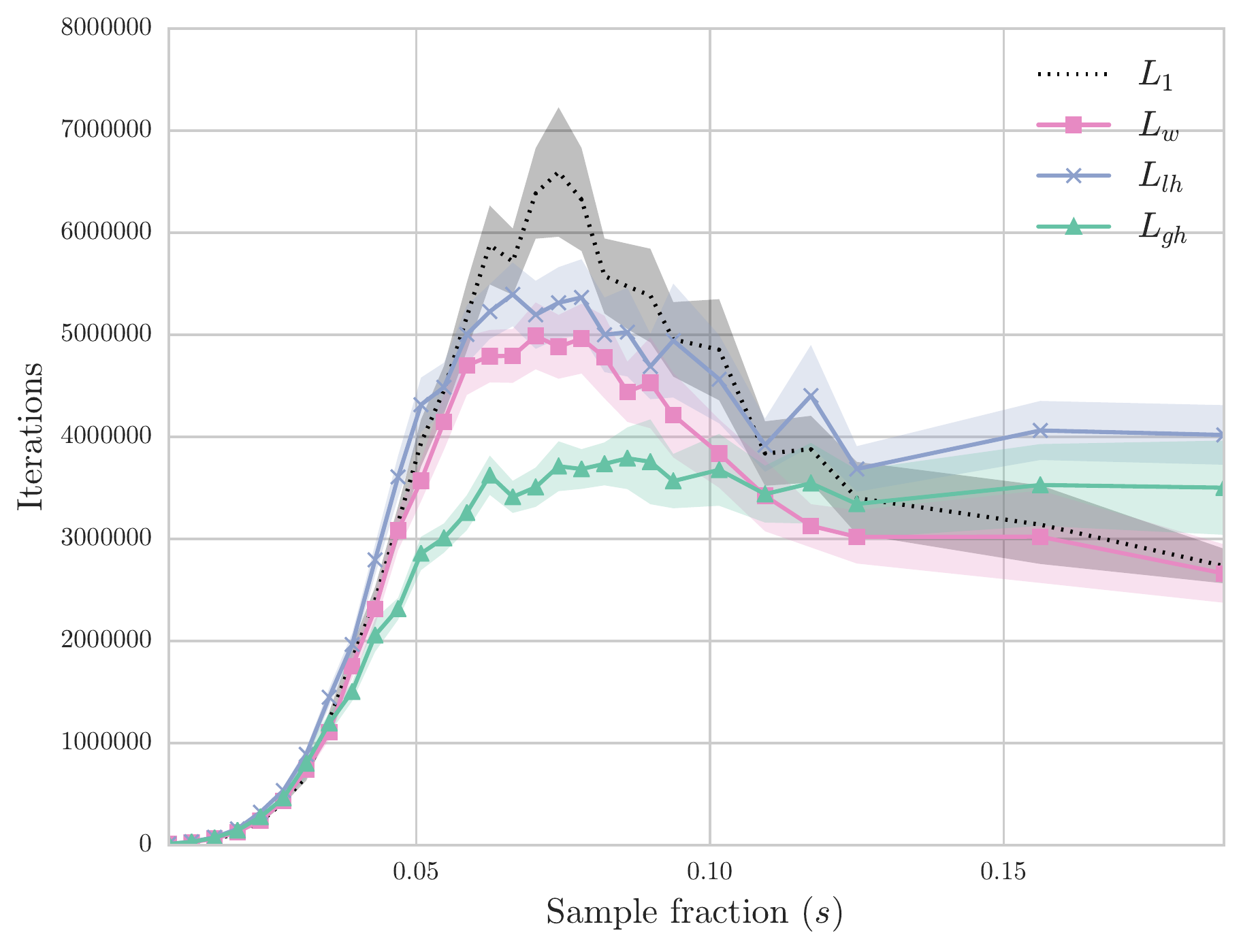}
        \caption{$5$-bit binary subtraction}
    \end{subfigure}\\
    \begin{subfigure}[t]{0.5\textwidth}
        \centering
        \includegraphics[width=0.9\linewidth]{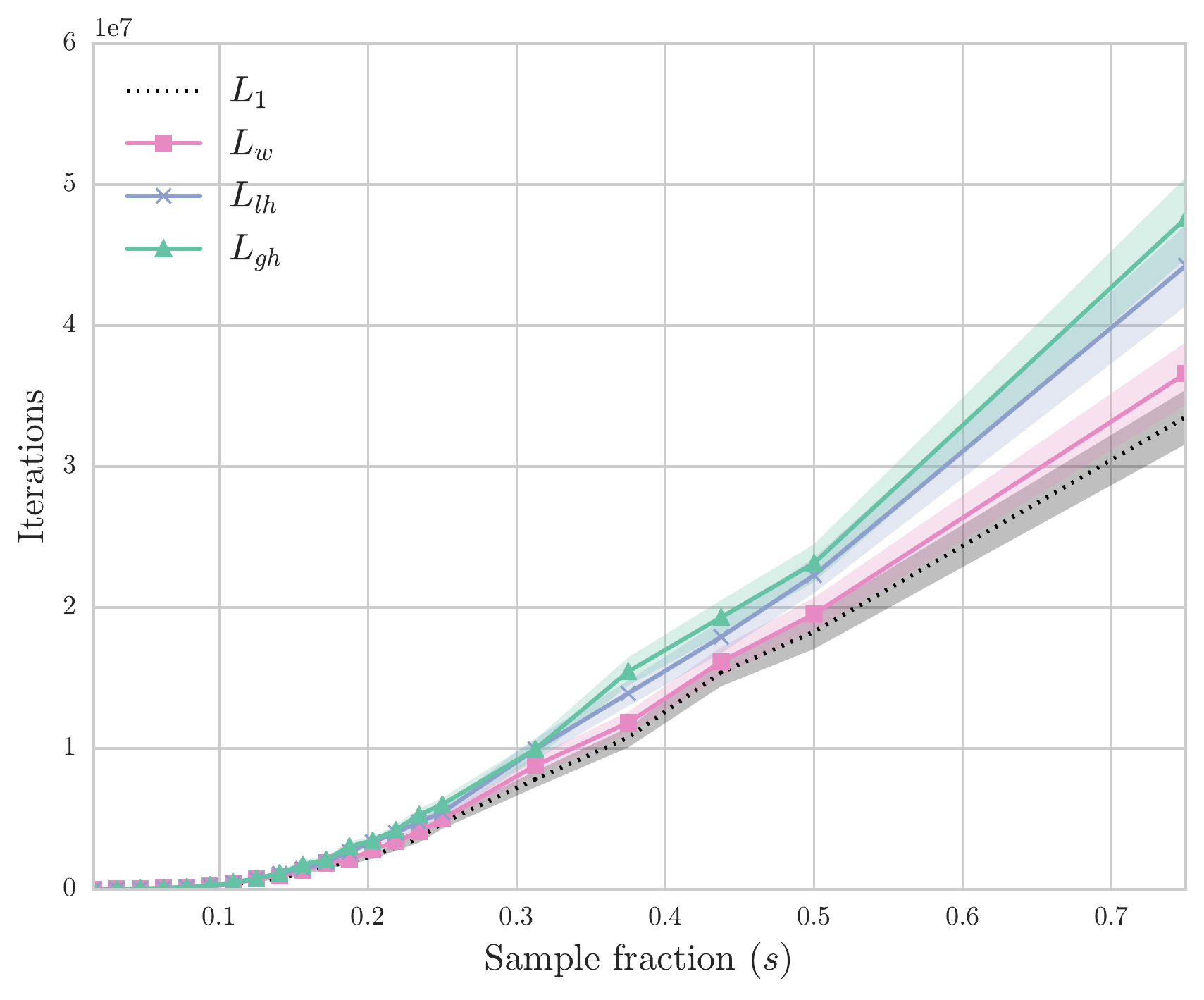}
        \caption{$9$-bit cascaded majority}
    \end{subfigure}%
    ~
    \begin{subfigure}[t]{0.5\textwidth}
        \centering
        \includegraphics[width=0.95\linewidth]{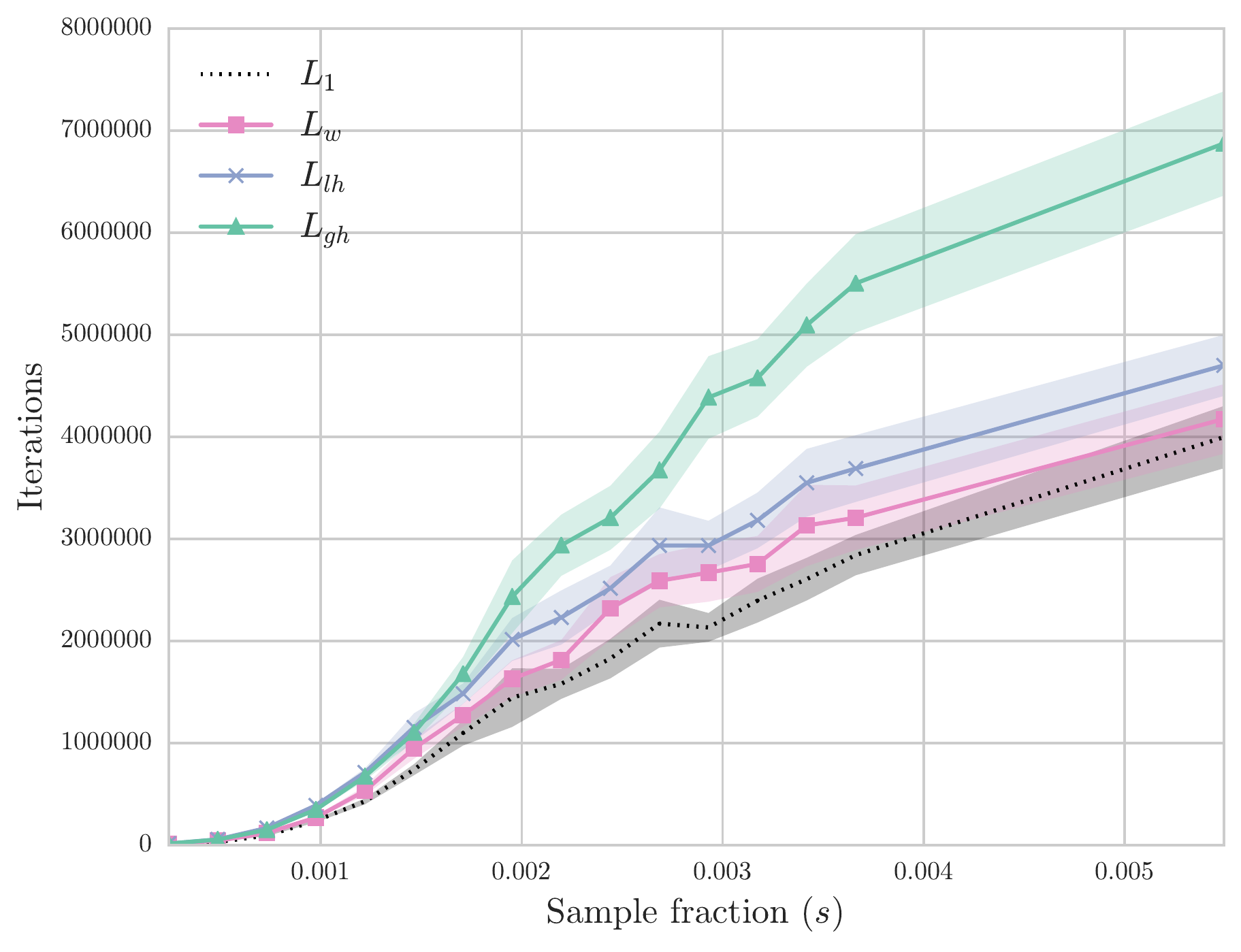}
        \caption{$15$-bit cascaded multiplexer}
    \end{subfigure}\\
    \begin{subfigure}[t]{0.5\textwidth}
        \centering
        \includegraphics[width=0.95\linewidth]{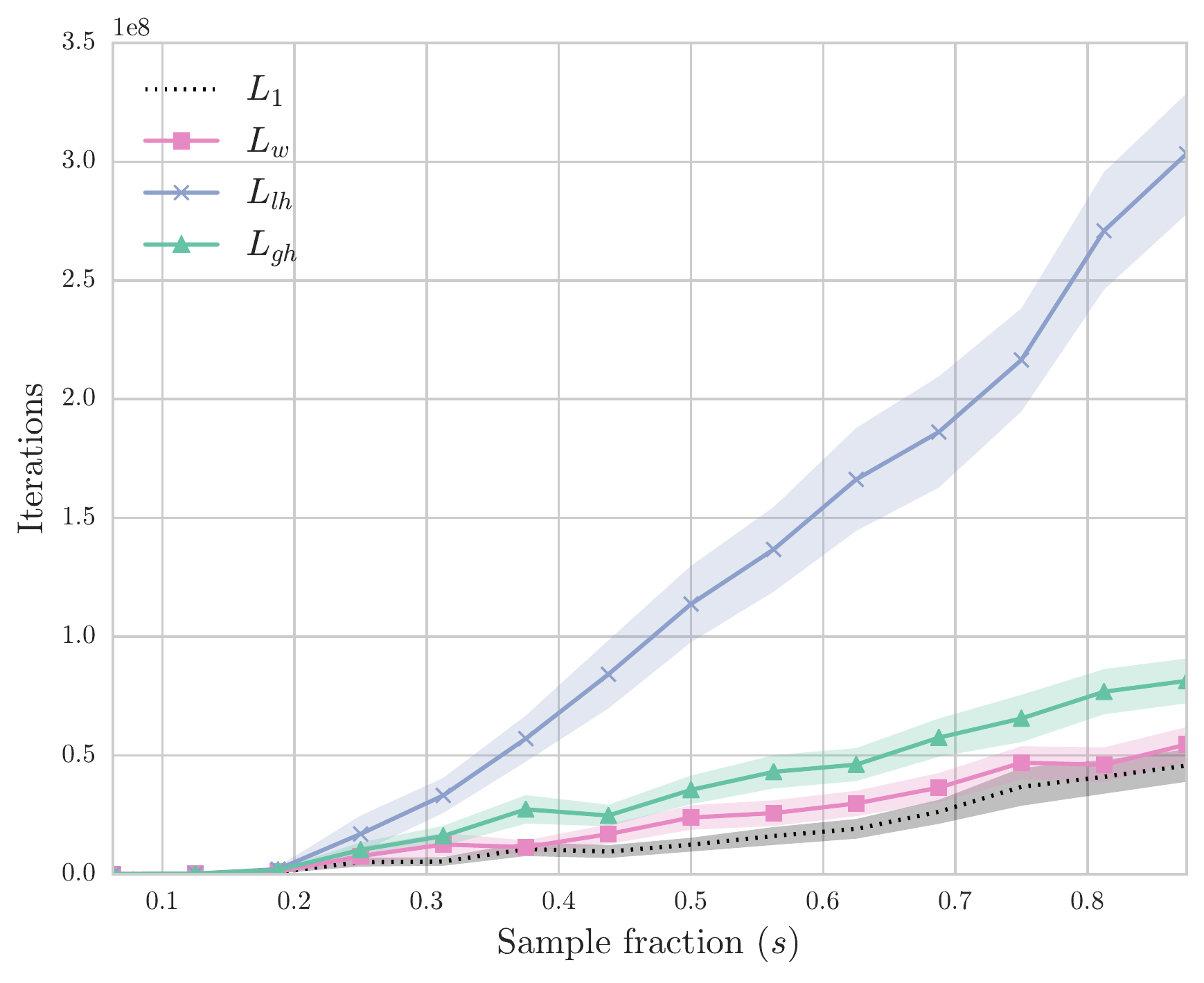}
        \caption{$7$-bit cascaded parity}
    \end{subfigure}\\
\caption{Mean number of iterations required to memorise the training set for networks trained using $L_1$, $L_{w}$, $L_{lh}$, and $L_{gh}$ on each test-bed problem. The variability of the impact on training speed is clear and there is no consistent trend across all problems.}
\label{fig:iterations}
\end{figure}

In this appendix we provide more detailed results. Figure~\ref{fig:add} shows the per-task training accuracy improvement (over $L_1$) for $3$-, $4$-, $5$-, and $6$-bit binary addition. Figure~\ref{fig:iterations} shows the training time (in iterations) as it varies with problem and loss function.

